\documentclass[letterpaper,twocolumn,10pt]{article}
\usepackage{float}
\usepackage[hmargin=0.75in,vmargin=1in,columnsep=0.33in]{geometry}
\usepackage[T1]{fontenc}
\usepackage[utf8]{inputenc}
\usepackage{mathptmx}
\usepackage{pslatex}
\usepackage{cite}

\usepackage{graphicx}
\usepackage{adjustbox}
\usepackage{booktabs}
\usepackage{amsmath,amssymb,amsthm,mathtools}
\usepackage{mathrsfs}
\usepackage{bbm}
\usepackage{makecell}
\usepackage{multirow}

\usepackage[most]{tcolorbox}
\usepackage{listings}
\usepackage{xcolor}
\usepackage[most]{tcolorbox}
\usepackage{listings}

\definecolor{promptbg}{RGB}{247,249,252}
\definecolor{promptframe}{RGB}{86,108,140}

\newtcblisting{promptbox}[1]{
  enhanced,
  breakable,
  colback=promptbg,
  colframe=promptframe,
  colbacktitle=promptframe!12,
  coltitle=black,
  title={#1},
  fonttitle=\bfseries\small,
  boxrule=0.6pt,
  arc=1mm,
  left=1mm,
  right=1mm,
  top=1mm,
  bottom=1mm,
  listing only,
  listing options={
    basicstyle=\ttfamily\small,
    breaklines=true,
    columns=fullflexible,
    keepspaces=true,
    showstringspaces=false,
    escapeinside={(*@}{@*)}
  }
}

\theoremstyle{remark}

\usepackage{amsmath}
\usepackage{amssymb}
\usepackage{amsthm}
\theoremstyle{plain}
\theoremstyle{remark}
\usepackage{xcolor}
\usepackage{listings}
\usepackage{algorithm}
\usepackage{algpseudocode}
\usepackage[kerning,spacing]{microtype}
\microtypecontext{spacing=nonfrench}
\usepackage{pifont}

\usepackage{tikz}
\usetikzlibrary{arrows.meta,positioning,fit,backgrounds,calc,shapes.geometric,%
                decorations.pathmorphing,shadows.blur,shadings}
\definecolor{wsacc}{HTML}{2F5597}    % weight-space accent (deep blue)
\definecolor{pxacc}{HTML}{B96A24}    % pixel-space accent (amber)
\definecolor{cardln}{HTML}{C7D0E0}   % card border
\definecolor{ink}{HTML}{1F2937}      % title text
\definecolor{subink}{HTML}{6B7280}   % subtitle text
\definecolor{gte}{HTML}{2A9D8F}      % layer group: te
\definecolor{gup}{HTML}{3F6CB5}      % layer group: up
\definecolor{gdown}{HTML}{8156A7}    % layer group: down
\definecolor{wsfill}{HTML}{EEF2FA}   \definecolor{pxfill}{HTML}{FBF1E6}
\definecolor{paper}{HTML}{FBFBFC}    \definecolor{hair}{HTML}{D7DCE3}
\definecolor{gtefill}{HTML}{E4F2F0}  \definecolor{gupfill}{HTML}{E7EDF7}
\definecolor{gdownfill}{HTML}{EFE9F4}\definecolor{chip}{HTML}{F4F6F9}
\usepackage{url}
\usepackage{hyperref}
\hypersetup{
  colorlinks=true,
  allcolors=blue!60!black,
  pdftitle={Discovering Undisclosed Concepts in LoRA Adapters for Diffusion-Based AIGC Models},
  pdfauthor={Yupeng Qiu, Han Fang, Dhananjai Sharma, Ee-Chien Chang}
}
\usepackage{xspace}

\newcommand{\sys}{\textsc{SCOPE}\xspace}

\makeatletter
\renewcommand{\section}{\@startsection{section}{1}{\z@}%
  {-3.5ex plus -1ex minus -.2ex}{2.3ex plus .2ex}%
  {\normalfont\large\bfseries}}
\renewenvironment{abstract}{%
  \begin{center}%
    {\large\bfseries\abstractname\vspace{-.5em}\vspace{\z@}}%
  \end{center}%
}{\par}
\makeatother

\date{}
\title{\Large\bfseries Discovering Undisclosed Concepts in LoRA Adapters for Diffusion-Based AIGC Models}

\author{
{\rm Yupeng Qiu}$^{1}$ \quad
{\rm Han Fang}$^{2}$\thanks{Corresponding author: Han Fang (\texttt{fanghan@ustc.edu.cn}).} \quad
{\rm Dhananjai Sharma}$^{1}$ \quad
{\rm Ee-Chien Chang}$^{1}$\\[0.3em]
{\rm\small $^{1}$National University of Singapore}\\
{\rm\small $^{2}$University of Science and Technology of China}\\[0.2em]
{\ttfamily\small qiu\_yupeng@u.nus.edu \quad fanghan@ustc.edu.cn}\\
{\ttfamily\small dhananjaisharma09@gmail.com \quad dcscec@nus.edu.sg}
}

\begin{document}
\maketitle

\begin{abstract}
Low-Rank Adaptation (LoRA) has made diffusion model personalization affordable and widely shareable, but it also creates moderation risks when harmful or problematic content is not faithfully disclosed. Hence, it is desired to develop a mechanism that, given a LoRA adapter, discovers the concept it was trained to generate, so as to support downstream investigation. We characterize the target concept through the conditioning input that most clearly exposes what the LoRA has learned. Such an input should induce coherent LoRA outputs that are distinctive from those of the base model and strongly align with the learned LoRA subspace. We therefore formulate concept discovery as searching for the conditioning input that jointly maximizes output coherence, LoRA-induced distinctiveness, and subspace alignment. Even with a clear objective, however, optimizing it over an open-ended input space is prohibitively expensive, since reliable evaluation of coherence and distinctiveness requires repeated diffusion rendering. Crucially, we observe that an input need not be rendered to determine whether it is promising. Its alignment with the learned LoRA subspace provides a lightweight signal for localizing promising candidates before visual verification. Based on this insight, we develop a two-stage framework for text-to-image LoRAs. Subspace-Guided Evolutionary Search (SGES) uses render-free subspace alignment to explore candidate prompts and retain a compact shortlist, while Paired Visual Identification (PVI) renders only these candidates and compares the LoRA and base-model outputs to identify those exhibiting coherent and distinctive generation behavior. This design enables efficient and reliable discovery of previously unknown concepts encoded in LoRA adapters.

\end{abstract}

\section{Introduction}

Diffusion models have substantially advanced text-to-image (T2I) generation~\cite{ho2020denoising,rombach2022high,saharia2022photorealistic,podell2024sdxl,croitoru2023diffusion,nichol2021glide,ramesh2022hierarchical}. Low-Rank Adaptation (LoRA)~\cite{hu2021lora} further reduces the cost of model customization and has become widely used for diffusion-model personalization~\cite{smith2023continual,wagner2026perpetuating}. By learning compact low-rank updates, LoRA can encode diverse visual concepts from only a small set of training images~\cite{kim2024datadream,gandikota2024concept}. Its low training and storage costs have facilitated the widespread creation and sharing of customized adapters~\cite{hu2021lora,wagner2026perpetuating}.

However, this accessibility also creates substantial moderation risks~\cite{carlini2023extracting,somepalli2023diffusion,shan2023glaze,van2023anti,schramowski2023safe,yang2024sneakyprompt}. LoRAs can encode copyrighted content, real-person identities, or NSFW concepts, and these capabilities may not be faithfully disclosed by their uploaders~\cite{wagner2026perpetuating,yuan2025lurks}. A large-scale study of CivitAI analyzed 221,304 model assets and found that 15.27\% (33,804) were flagged as targeting real individuals~\cite{wagner2026perpetuating}. Regulatory scrutiny has also increased. For example, the U.S. TAKE IT DOWN Act addresses non-consensual intimate imagery, including AI-generated depictions~\cite{takeitdown2025}, while the EU AI Act imposes transparency requirements on deepfakes and other AI-generated content~\cite{euaiact2024}. Yet moderation on model-sharing platforms still relies heavily on uploader-provided descriptions, tags, trigger words, and sample images~\cite{wagner2026perpetuating}; CivitAI and Hugging Face illustrate the model-sharing setting~\cite{civitai2026models,huggingface2026modelhub}. Such metadata becomes unreliable when disclosure is incomplete or misleading~\cite{yuan2025lurks}. Hence, it is desirable to develop a mechanism that, given a LoRA adapter, discovers the concept it was trained to generate.

Existing techniques provide only partial solutions to this problem. Difference-discovery methods identify changes between a customized model and its base model~\cite{laria2026driftscope,dunlap2025discovering,dunlap2024describing}, but LoRA adaptation often alters not only the primary training target but also related or collateral content~\cite{laria2026driftscope}. As a result, detecting what has changed does not reveal which change best represents what the adapter has learned. Concept-verification methods take the opposite route: they test whether a predefined concept is encoded by the LoRA, either through prompt recovery~\cite{wen2023hard} or direct model comparison~\cite{yuan2025lurks}. Their effectiveness, however, is bounded by the concepts specified in advance. Both directions therefore leave the same fundamental question unresolved: \textbf{when the hidden concept is unknown, what exactly should LoRA forensics seek to discover?}

To answer this question, we characterize the hidden concept through the conditioning input that most clearly exposes what the LoRA has learned. A revealing input should induce a coherent and meaningful output distribution under the LoRA, make this distribution clearly distinctive from that of the base model, and be strongly represented in the learned low-rank subspace.
These criteria capture complementary evidence from generation behavior and model parameters: the former establishes whether a stable LoRA-specific behavior is revealed, while the latter indicates whether that behavior is strongly supported by the adapter itself.
We therefore formulate hidden-concept discovery as searching for the conditioning input that jointly maximizes output coherence, LoRA-induced distinctiveness, and LoRA-subspace alignment.

However, even with this target well defined, solving the resulting optimization is far from straightforward. The search space of possible conditioning inputs is vast and open-ended, while evaluating output coherence and LoRA-induced distinctiveness requires repeated diffusion rendering. Crucially, we find a strong correspondence between parameter-space alignment and generation-space evidence: inputs that align more strongly with the learned LoRA subspace are substantially more likely to induce coherent and distinctive LoRA-specific behavior after rendering. This finding provides a render-free signal for navigating the large input space, allowing expensive visual evaluation to be concentrated on only a small set of promising candidates.

Building on this finding, we develop a two-stage framework for text-to-image LoRA forensics. We instantiate the conditioning input as a textual prompt and first use \textbf{Subspace-Guided Evolutionary Search} \textbf{(SGES)} to explore the open-ended prompt space using LoRA-subspace alignment alone, without diffusion rendering. SGES progressively concentrates the search on highly aligned prompts and produces a compact candidate set. \textbf{Paired Visual Identification (PVI)} then renders only these candidates with both the LoRA and base model and identifies those that exhibit coherent LoRA outputs and clear LoRA-induced distinctiveness. In this way, large-scale exploration is performed with a lightweight parameter-space signal, while expensive generation is reserved for final visual identification.

The main contributions of this work are as follows:
\begin{itemize}

\item  We study the important problem of discovering previously unknown concepts encoded in LoRA adapters and identify three essential characteristics of a valid forensic target: output coherence, LoRA-induced distinctiveness, and LoRA-subspace alignment.
These characteristics provide a principled basis for defining and quantifying hidden-concept discovery.

\item We uncover a strong correspondence between LoRA-subspace alignment and LoRA-specific generation behavior, and exploit this finding to develop SGES and PVI.
The resulting framework performs large-scale search without diffusion rendering and reserves expensive visual evaluation for only a compact set of candidates.

\item We construct a benchmark covering diverse concept types and scenarios, and compare against representative difference-discovery, prompt-recovery, and concept-verification methods. We further introduce a distributional metric to evaluate how faithfully the discovered result captures the hidden content encoded by the LoRA.

\end{itemize}

\section{Background and Related Work}
\subsection{Diffusion Models}

Diffusion models generate images by learning to reverse a gradual stochastic noising process~\cite{sohl2015deep,croitoru2023diffusion}. Denoising Diffusion Probabilistic Models (DDPMs) introduced an effective noise-prediction objective~\cite{ho2020denoising}, while subsequent work incorporated text conditioning and latent-space denoising, enabling efficient and high-fidelity text-to-image generation~\cite{rombach2022high,saharia2022photorealistic,nichol2021glide,ramesh2022hierarchical}.
\\
\textbf{Diffusion process.}
Let $\mathbf{x}_0$ denote a clean image or its latent representation. The forward process gradually adds Gaussian noise according to~\cite{sohl2015deep,ho2020denoising}
\begin{equation}
q(\mathbf{x}_t\mid\mathbf{x}_{t-1})
=
\mathcal N\left(
\sqrt{\alpha_t}\mathbf{x}_{t-1},
(1-\alpha_t)\mathbf I
\right),
\end{equation}
where $\alpha_t=1-\beta_t$ and $\beta_t$ controls the noise level.
Defining $\bar{\alpha}_t=\prod_{s=1}^{t}\alpha_s$, a noisy sample at
timestep $t$ can be obtained directly as
\begin{equation}
\mathbf{x}_t
=
\sqrt{\bar{\alpha}_t}\mathbf{x}_0
+
\sqrt{1-\bar{\alpha}_t}\boldsymbol{\epsilon},
\qquad
\boldsymbol{\epsilon}\sim\mathcal N(\mathbf 0,\mathbf I).
\end{equation}
A denoising network learns to predict the added noise and iteratively reverses this process from Gaussian noise $\mathbf{x}_T$ to $\mathbf{x}_0$. Modern T2I models commonly perform denoising in a compressed latent space and decode the final latent into an image~\cite{rombach2022high}.
\\
\textbf{Text conditioning.}
For a text input $x$, a text encoder produces contextual token representations
$
\mathbf C(x)
\in
\mathbb R^{d_c\times L(x)},
$
where $L(x)$ is the token length and $d_c$ is the representation dimension. These representations condition the denoising network through cross-attention~\cite{vaswani2017attention,rombach2022high}. Let $\mathbf F_t$ denote intermediate visual features at timestep $t$. The query, key, and value projections are
\begin{equation}
\mathbf Q=\mathbf W_Q\mathbf F_t,\qquad
\mathbf K=\mathbf W_K\mathbf C(x),\qquad
\mathbf V=\mathbf W_V\mathbf C(x).
\end{equation}
The query therefore depends on the evolving visual features, whereas the key and value inputs are determined directly by the text conditioning. This mechanism makes generation sensitive to both the
tokens in the text input and their context.

With text conditioning, the denoising network is trained using the noise-prediction objective~\cite{ho2020denoising,rombach2022high}
\begin{equation}
\mathcal L_{\mathrm{DM}}
=
\mathbb E_{\mathbf{x}_0,x,t,\boldsymbol{\epsilon}}
\left[
\left\|
\boldsymbol{\epsilon}
-
\boldsymbol{\epsilon}_{\theta}
\bigl(\mathbf{x}_t,t,\mathbf C(x)\bigr)
\right\|_2^2
\right].
\end{equation}

At inference time, classifier-free guidance~\cite{ho2022classifier} can further strengthen text conditioning by combining conditional and unconditional noise predictions:
\begin{equation}
\hat{\boldsymbol{\epsilon}}_{\theta}
=
\boldsymbol{\epsilon}_{\theta}
(\mathbf{x}_t,t,\varnothing)
+
\eta
\left[
\boldsymbol{\epsilon}_{\theta}
\bigl(\mathbf{x}_t,t,\mathbf C(x)\bigr)
-
\boldsymbol{\epsilon}_{\theta}
(\mathbf{x}_t,t,\varnothing)
\right],
\end{equation}
where $\varnothing$ denotes the empty condition and $\eta$ is the guidance scale.

\subsection{Low-Rank Adaptation}
Parameter-efficient fine-tuning (PEFT) adapts large pretrained models while updating only a small number of parameters~\cite{houlsby2019parameter,ding2023parameter,li2021prefix,lester2021power,zaken2022bitfit,liu2022few,xie2023difffit}. Among PEFT methods, Low-Rank Adaptation (LoRA)~\cite{hu2021lora} has become widely used for diffusion-model personalization~\cite{smith2023continual,wagner2026perpetuating}. LoRA keeps the base model fixed and learns compact low-rank weight updates, enabling visual concepts to be learned from relatively small training sets with low training and storage costs~\cite{kim2024datadream,gandikota2024concept}.
\\
\textbf{Low-rank weight adaptation.}
Consider a pretrained weight matrix $\mathbf W_0\in\mathbb R^{d_{\mathrm{out}}\times d_{\mathrm{in}}}$. LoRA freezes $\mathbf W_0$ and represents its learned update as $\Delta\mathbf W=\mathbf B\mathbf A$, where $\mathbf A\in\mathbb R^{r\times d_{\mathrm{in}}}$, $\mathbf B\in\mathbb R^{d_{\mathrm{out}}\times r}$, and $r\ll\min(d_{\mathrm{in}},d_{\mathrm{out}})$~\cite{hu2021lora}. For an input $\mathbf z\in\mathbb R^{d_{\mathrm{in}}}$, the adapted layer computes
\begin{equation}
\mathbf h
=
\mathbf W_0\mathbf z+\Delta\mathbf W\mathbf z,
\qquad
\Delta\mathbf W=\mathbf B\mathbf A.
\label{eq:lora-update}
\end{equation}
Only $\mathbf A$ and $\mathbf B$ are optimized, reducing the number of trainable parameters from $d_{\mathrm{in}}d_{\mathrm{out}}$ to $r(d_{\mathrm{in}}+d_{\mathrm{out}})$.

In T2I diffusion models, LoRA modules can be inserted into selected linear projections in the text encoder and denoising network~\cite{kim2024datadream}.
The learned updates therefore modify text representations, denoising behavior, or their interactions, depending on the adapted sites. After training, the resulting LoRA weights can be stored and shared independently of the base model~\cite{hu2021lora}.
\\
\textbf{Input-space interpretation.} The change introduced by a LoRA update is $\Delta\mathbf h=\Delta\mathbf W\mathbf z$. Let $\Delta\mathbf W=\mathbf U\boldsymbol{\Sigma}\mathbf V^\top$ denote its compact singular value decomposition, with rank $\rho$, right singular vectors $\{\mathbf v_j\}_{j=1}^{\rho}$, and singular values $\{\sigma_j\}_{j=1}^{\rho}$. The magnitude of the LoRA response satisfies 
\begin{equation}
\|\Delta\mathbf h\|_2^2 = \|\Delta\mathbf W\mathbf z\|_2^2 = \sum_{j=1}^{\rho} \sigma_j^2 (\mathbf v_j^\top\mathbf z)^2.
\end{equation}
Thus, the response largely depends on the alignment of the input with each right singular vector. Inputs orthogonal to the right singular subspace produce no LoRA response. The leading right singular vectors therefore define the dominant input subspace of the LoRA update, motivating our subsequent parameter-space design.

\subsection{Content Moderation for LoRA Adapters}
\textbf{Metadata-based review.}
Model-sharing platforms primarily rely on uploader-provided descriptions, tags, trigger words, and sample images to review shared adapters~\cite{wagner2026perpetuating}.
However, such metadata cannot reliably reveal encoded content when disclosure is incomplete or misleading~\cite{yuan2025lurks}.
\\
\textbf{Concept-agnostic difference discovery.}
Related methods discover differences between models or their generated outputs without specifying a concept in advance. VisDiff~\cite{dunlap2024describing} describes semantic differences between image sets, CompCon~\cite{dunlap2025discovering} searches for prompts that expose differences between T2I models, and DriftScope~\cite{laria2026driftscope} identifies concept changes caused by model adaptation. Discrete prompt optimization~\cite{wang2024discrete} can likewise be adapted to search for LoRA--base differences. However, adaptation may also affect collateral concepts~\cite{laria2026driftscope}; therefore, identifying what changes does not determine which change best represents the creator-intended concept. Output-based discovery can also require extensive diffusion rendering during search~\cite{dunlap2025discovering,wang2024discrete}.
\\
\textbf{Concept-specified verification.}
A complementary direction starts from a concept specified in advance. PEZ~\cite{wen2023hard} recovers prompts for a given concept, providing a way to test whether that concept can be elicited from the LoRA model.
PAIA~\cite{yuan2025lurks} more directly verifies whether a fine-tuned diffusion model has learned a specified concept by comparing its denoising behavior with that of the base model. While such verification can be efficient for an individual concept, its coverage is limited to concepts provided beforehand. Discovering an unknown concept therefore requires repeated verification over a potentially large candidate space.
\\
\textbf{Summary.}
Existing methods therefore either discover LoRA-specific changes without identifying the creator-intended one, or verify concepts that must already be specified. In contrast, our work formulates the creator-intended concept as the \emph{dominant hidden concept} and discovers it jointly from LoRA parameter space and generation behavior.

\section{Problem Formulation}
\label{sec:problem-formulation}
\subsection{Problem Setting}
\label{sec:problem-setting}

Suppose we are given a LoRA adapter $\mathcal A$ and its compatible pretrained T2I base model $\mathcal B$. Loading $\mathcal A$ into $\mathcal B$ yields the LoRA-enabled model $\mathcal B_{\mathcal A}$. 
The adapter may affect multiple concepts. We define the \emph{dominant hidden concept} as the creator-intended primary training target that is repeatedly reinforced during adaptation, and distinguish it from collateral concepts affected incidentally~\cite{laria2026driftscope}. The moderator has access only to $\mathcal A$ and $\mathcal B$, without reliable metadata, known trigger words, predefined concept queries, reference images, or training data. The goal is to discover a \emph{carrier prompt} $p^\star$ that reveals the dominant hidden concept encoded by $\mathcal A$.
\\
\textbf{Prompt-conditioned output distributions.}
Let $\mathcal X$ denote the image space and let $\mathcal E:\mathcal X\rightarrow\mathbb R^q$ be a fixed semantic image encoder~\cite{radford2021learning}. We use its normalized representation
% $
\begin{equation}
\phi(\mathbf x)
=
\frac{\mathcal E(\mathbf x)}
{\|\mathcal E(\mathbf x)\|_2}.
\label{eq:normalized-semantic-encoder}
% $
\end{equation}

Let $\mathcal P$ denote the prompt space and let $Z\sim\pi_Z$ represent the randomness in diffusion generation. For a prompt $p\in\mathcal P$, the semantic output of the LoRA-enabled model satisfies
% $
\begin{equation}
\mathbf Y_{\mathcal A}(p)
=
\phi\!\left(
\mathcal B_{\mathcal A}(p,Z)
\right)
\sim
Q_{\mathcal A}(\cdot\mid p),
\label{eq:lora-conditional-semantic-output}
\end{equation}
% $
where $Q_{\mathcal A}(\cdot\mid p)$ denotes the prompt-conditioned semantic distribution induced by $\mathcal B_{\mathcal A}$ over diffusion randomness. Similarly,
% $
\begin{equation}
\mathbf Y_{\mathcal B}(p)
=
\phi\!\left(
\mathcal B(p,Z)
\right)
\sim
Q_{\mathcal B}(\cdot\mid p),
\label{eq:base-conditional-semantic-output}
\end{equation}
% $
defines the corresponding prompt-conditioned semantic distribution of the compatible base model.

\subsection{Dominant Hidden Concept Formulation}
\label{sec:hidden-concept-formulation}
We characterize the dominant hidden concept from two complementary views: generation behavior and LoRA parameter space. A prompt revealing this concept should induce coherent LoRA outputs that are distinctive from those of the compatible base model and align strongly with the dominant input subspaces of the learned LoRA updates.
We formalize these requirements as output coherence, LoRA-induced distinctiveness, and LoRA-subspace alignment.
\\
\textbf{Output coherence.}
A prompt revealing the dominant hidden concept should induce a coherent output distribution, with little variation across diffusion samples generated from the same prompt. Let $\mathcal C(\cdot)$ denote a coherence functional over output distributions, with larger values indicating stronger coherence. We define the output coherence of prompt $p$ as
\begin{equation}
\operatorname{Coh}_{\mathcal A}(p)
=
\mathcal C\!\left(
Q_{\mathcal A}(\cdot\mid p)
\right).
\label{eq:output-coherence}
\end{equation}
\textbf{LoRA-induced distinctiveness.}
Output coherence alone is insufficient, since a prompt may consistently produce a common concept that is already supported by the compatible base model. A prompt revealing the dominant hidden concept should therefore induce an output distribution distinct from that of the base model.
Let $\mathcal D(\cdot,\cdot)$ denote a discrepancy functional between output distributions, with larger values indicating greater discrepancy. We define the LoRA-induced distinctiveness of prompt $p$ as
\begin{equation}
\operatorname{Dist}_{\mathcal A,\mathcal B}(p)
=
\mathcal D\!\left(
Q_{\mathcal A}(\cdot\mid p),
Q_{\mathcal B}(\cdot\mid p)
\right).
\label{eq:lora-induced-distinctiveness}
\end{equation}
\textbf{LoRA-subspace alignment.}
Output coherence and LoRA-induced distinctiveness are still insufficient, since adaptation may also affect collateral concepts~\cite{laria2026driftscope}. We assume that the primary training target is shared across concept-specific examples and is therefore expected to leave stronger evidence in the leading input directions than incidental content. A revealing prompt should consequently align strongly with these directions. 
Let $\mathcal S_{\mathcal A}$ denote the input subspaces spanned by the leading right singular directions of the LoRA updates, and let $\mathcal H(\cdot,\cdot)$ denote a prompt--subspace alignment functional, with larger values indicating stronger alignment. We define
\begin{equation}
\operatorname{Align}_{\mathcal A}(p)
=
\mathcal H\!\left(
p,\mathcal S_{\mathcal A}
\right).
\label{eq:lora-subspace-alignment}
\end{equation}
\textbf{Dominant hidden concept.}
The three properties jointly characterize a prompt that reveals the dominant hidden concept. We define the dominant carrier prompt as
\begin{equation}
p^\star
\in
\operatorname*{arg\,max}_{p\in\mathcal P}
\Big[
\alpha\,
\operatorname{Coh}_{\mathcal A}(p)
+
\beta\,
\operatorname{Dist}_{\mathcal A,\mathcal B}(p)
+
\gamma\,
\operatorname{Align}_{\mathcal A}(p)
\Big],
\label{eq:dominant-hidden-concept}
\end{equation}
where $\alpha,\beta,\gamma\geq0$ balance the three components. The output distribution induced by $p^\star$ represents the discovered dominant hidden concept, while $p^\star$ serves as its carrier prompt.

\subsection{Target-Concept Discrepancy}
\label{sec:target-concept-discrepancy}

The formulation above defines the dominant hidden concept but does not
quantify how faithfully a discovered carrier prompt reveals a given target
concept. To provide such a measure, let $c$ denote a target concept and
$P_c^{\mathrm{ref}}$ its reference image distribution.
For
$X_c^{\mathrm{ref}}\sim P_c^{\mathrm{ref}}$, let
\begin{equation}
\mathbf Y_c^{\mathrm{ref}}
=
\phi(X_c^{\mathrm{ref}})
\sim
Q_c^{\mathrm{ref}},
\label{eq:target-concept-distribution}
\end{equation}
where $Q_c^{\mathrm{ref}}$ is the corresponding distribution in the
semantic representation space.
We represent the target concept by the mean direction of its reference
distribution:
\begin{equation}
\boldsymbol{\mu}_c^{\mathrm{ref}}
=
\mathbb E_{\mathbf Y\sim Q_c^{\mathrm{ref}}}[\mathbf Y],
\qquad
\overline{\boldsymbol{\mu}}_c^{\mathrm{ref}}
=
\frac{\boldsymbol{\mu}_c^{\mathrm{ref}}}
{\|\boldsymbol{\mu}_c^{\mathrm{ref}}\|_2}.
\label{eq:target-concept-direction}
\end{equation}
The target-concept discrepancy of the carrier prompt $p^\star$ is then
defined as
\begin{equation}
\Delta_{\mathcal A}(p^\star;c)
=
1-
\mathbb E_{\mathbf Y\sim Q_{\mathcal A}(\cdot\mid p^\star)}
\left[
\mathbf Y^\top
\overline{\boldsymbol{\mu}}_c^{\mathrm{ref}}
\right].
\label{eq:target-concept-discrepancy}
\end{equation}
Since $\phi(\cdot)$ is $\ell_2$-normalized,
Equation~\eqref{eq:target-concept-discrepancy} measures the expected
cosine distance between the LoRA-enabled outputs induced by
$p^\star$ and the semantic direction of the target concept $c$.
A smaller $\Delta_{\mathcal A}(p^\star;c)$ therefore indicates that
the discovered carrier prompt more faithfully reveals the target concept.

In evaluation, we estimate the reference direction from a finite set of
reference images and average the cosine distance over generated outputs.
This quantity is used only for evaluation and is unavailable during
hidden-concept discovery.

\subsection{Threat Model}
\label{sec:threat-model}
We consider a model-sharing platform on which a creator uploads a LoRA adapter whose dominant hidden concept is not reliably disclosed~\cite{civitai2026models,huggingface2026modelhub}.
\\
\textbf{Creator capabilities.}
The creator controls the LoRA training process and may omit or obscure information about the concept encoded by the adapter, including its description, trigger words, and example images.\\
\textbf{Platform capabilities.}
The platform moderator has white-box access to the uploaded adapter $\mathcal A$ and its compatible base model $\mathcal B$. The moderator may inspect the LoRA parameters and query both $\mathcal B_{\mathcal A}$ and $\mathcal B$, but has no access to the training data, known trigger words, predefined concept queries, or reference images during discovery.\\
\textbf{Moderator goal.}
The moderator aims to recover a carrier prompt $p^\star$ that satisfies the three desiderata formalized in Equation~\eqref{eq:dominant-hidden-concept}, using only information available from $\mathcal A$ and $\mathcal B$.

\section{Methodology}
\label{sec:methodology}

\subsection{Framework Overview}
\label{sec:framework-overview}

\begin{figure*}[t]
    \centering
    \includegraphics[width=\linewidth]{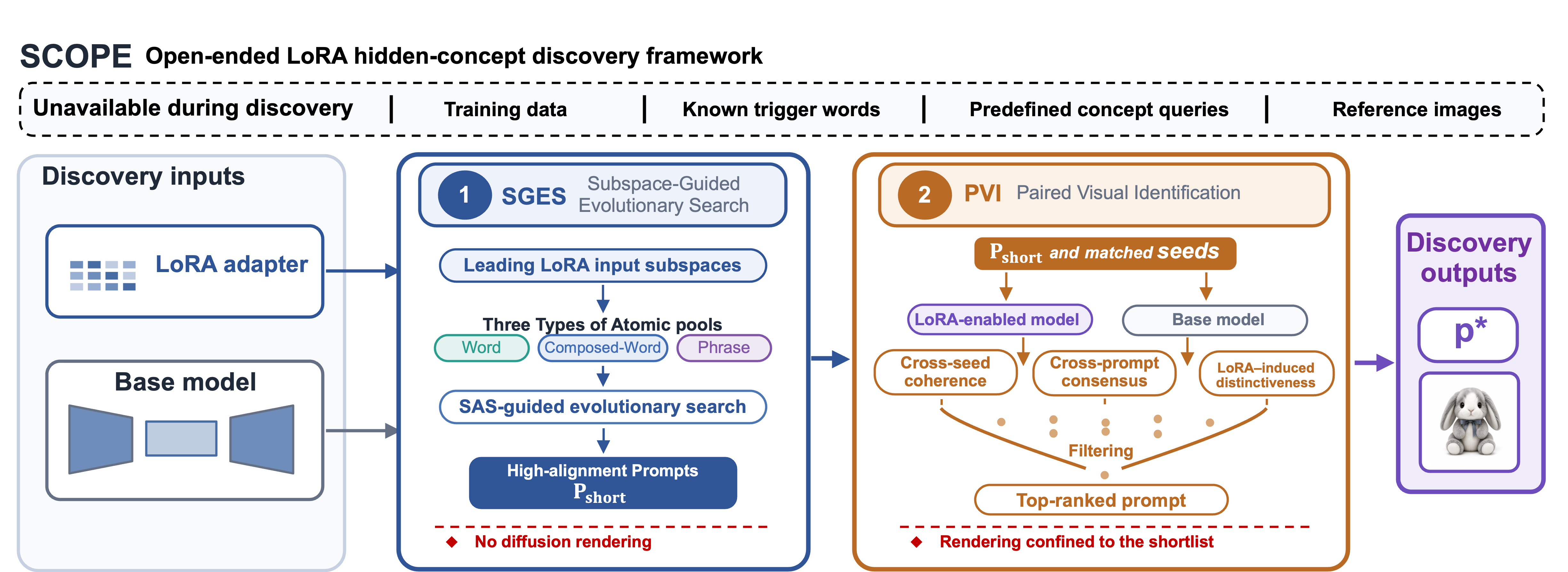}
    \caption{End-to-end \sys workflow. Training data, trigger
    words, predefined concept queries, and reference images are unavailable
    during discovery. Given only a LoRA adapter and its compatible base model,
    SGES extracts
    leading LoRA input subspaces, constructs Atomic pools, and performs
    SAS-guided evolutionary search to produce $\mathcal P_{\mathrm{short}}$
    without diffusion rendering. PVI renders the shortlisted prompts
    with the LoRA-enabled and base models under matched seeds, filters and ranks
    them using coherence, cross-prompt consensus, and LoRA-induced
    distinctiveness, and returns the top-ranked prompt $p^\star$ with
    its LoRA-enabled generated images.}
    \label{fig:scope-overview}
\end{figure*}

Equation~\eqref{eq:dominant-hidden-concept} characterizes the carrier prompt through output coherence, LoRA-induced distinctiveness, and LoRA-subspace alignment. Directly optimizing these properties over the prompt space is impractical: the prompt space is combinatorially large, while evaluating the two output-side properties requires repeated diffusion rendering.
We therefore develop \sys, a two-stage framework that separates render-free prompt search from output-side identification, as illustrated in Figure~\ref{fig:scope-overview}. SGES first uses an empirical alignment score to localize a high-alignment shortlist without diffusion rendering. PVI then renders only that shortlist, rejects candidates whose LoRA-enabled outputs are less coherent than the corresponding outputs from the compatible base model, and ranks the survivors using cross-prompt consensus together with an empirical distinctiveness score. The highest-ranked candidate is returned as the dominant carrier prompt $p^\star$.

\subsection{Subspace-Guided Evolutionary Search}
\label{sec:sges}
SGES first addresses the LoRA-subspace alignment term in Equation~\eqref{eq:dominant-hidden-concept}. Unlike the two output-side properties, this alignment can be evaluated from the LoRA parameters and text-conditioned representations without diffusion rendering. SGES uses this signal to reduce the combinatorial prompt space to a
compact set of high-alignment candidates for subsequent visual identification.
\\
\textbf{Text-conditioning sites.}
To obtain a render-free alignment signal, SGES focuses on LoRA sites whose inputs are determined by text conditioning and can be computed without executing the diffusion denoising trajectory. We consider LoRA sites in the text encoder and the key/value projections of text-facing cross-attention layers in the denoising backbone~\cite{rombach2022high,podell2024sdxl}.
For the U-Net-based models studied in this work, we group the retained sites into text-encoder $\mathrm{te}$, down-path $\mathrm{down}$, and up-path $\mathrm{up}$ groups. Let $\mathcal J_b$ denote the retained LoRA sites in group $b$, and define the groups present in adapter $\mathcal A$ as

\begin{equation}
\mathcal G_{\mathcal A}
=
\left\{
b\in\{\mathrm{te},\mathrm{down},\mathrm{up}\}
:
|\mathcal J_b|>0
\right\}.
\label{eq:site-groups}
\end{equation}
denote the three site groups, and let $\mathcal J_b$ denote the retained LoRA sites in group $b\in\mathcal G_{\mathcal A}$. Architecture-specific site selection and grouping details are provided in Appendix~\ref{app:sges-sites}.

For a text input $x$ and a retained site $s\in\mathcal J_b$, let
\begin{equation}
\mathbf H_s(x)
=
\left[
\mathbf h_{s,1}(x),
\ldots,
\mathbf h_{s,L(x)}(x)
\right]
\in
\mathbb R^{d_s^{\mathrm{in}}\times L(x)}
\label{eq:site-contextual-representation}
\end{equation}
denote the text-conditioned input sequence at site $s$, where $L(x)$ is the tokenized sequence length and $u$ indexes a token position. We denote the $\ell_2$-normalized representation at position $u$ by
$\widehat{\mathbf h}_{s,u}(x)$.
\\
\textbf{Dominant input subspaces.}
For each retained LoRA site $s$, the learned weight update can be written as
$
\Delta\mathbf W_s
=
\mathbf B_s\mathbf A_s
=
\mathbf U_s
\boldsymbol\Sigma_s
\mathbf V_s^\top.
$
For an input representation $\mathbf h$, the response $\Delta\mathbf W_s\mathbf h$ is determined by the projections of $\mathbf h$ onto the right singular vectors of $\Delta\mathbf W_s$. Accordingly, the right singular vectors associated with the largest singular values therefore define the dominant input subspace of the LoRA update. We retain the top-$k$ directions $\{\mathbf v_{s,j}\}_{j=1}^{k}$ as the dominant input subspace at site $s$.
\\
\textbf{Site-position alignment.}
Given the dominant input directions at site $s$, we measure the alignment of each token position with these directions.

\begin{equation}
g_{s,u}(x)
=
\prod_{j=1}^{k}
\left[
\left(
\mathbf v_{s,j}^{\top}
\widehat{\mathbf h}_{s,u}(x)
\right)^2
\right].
\label{eq:site-position-response}
\end{equation}
The product favors token positions that align consistently across the retained dominant directions, rather than exhibiting a strong projection onto only one direction.
\\
\textbf{Composed-word discovery.}
The tokenizer vocabulary consists of lexical units that may represent only fragments of a word~\cite{sennrich2016neural,radford2021learning}. A candidate word may therefore span multiple tokenizer units. SGES performs composed-word discovery separately for each site group, combining promising units into candidate words before prompt-level search.

For a candidate word $w$ composed of one to three lexical units, let $L(w)$ denote its tokenized sequence length. Because different LoRA sites may respond most strongly at different positions, we identify the position jointly supported by the sites in group $b$:
\begin{equation}
u_b^\star(w)
=
\operatorname*{arg\,max}_{1\leq u\leq L(w)}
\left[
\prod_{s\in\mathcal J_b}
g_{s,u}(w)
\right]^{1/|\mathcal J_b|}.
\label{eq:shared-position}
\end{equation}
We then score the candidate by
\begin{equation}
S_b(w)
=
\frac{1}{|\mathcal J_b|}
\sum_{s\in\mathcal J_b}
g_{s,u_b^\star(w)}(w).
\label{eq:composition-score}
\end{equation} 
This selection favors positions whose alignment is consistently supported across LoRA sites in the same group. SGES progressively extends high-scoring lexical units into two- and three-unit candidate words; detailed construction rules are provided in
Appendix~\ref{app:bpe-beam}.\\
\textbf{Atomic candidate pools.}
SGES forms its prompt-search units from vocabulary words, the composed words discovered above, and common phrases. We refer to each such unit as an Atomic. An Atomic may span multiple tokenizer positions but is treated as a single unit during prompt search.

For an Atomic $a$ occurring in the text context $x$, let $\operatorname{pos}_x(a)$ denote its token positions. Its response in site group $b$ is defined as
\begin{equation}
R_b(a\mid x)
=
\frac{1}{|\mathcal J_b|}
\sum_{s\in\mathcal J_b}
\max_{u\in\operatorname{pos}_x(a)}
g_{s,u}(x).
\label{eq:atomic-response}
\end{equation}
Each LoRA site therefore contributes the position within $a$ that aligns most strongly with its dominant input subspace.

To construct compact search pools, each Atomic is first evaluated in isolation using $R_b(a\mid a)$ and assigned to the site group in which
it ranks highest. Each group retains its top-ranked Atomics, forming the candidate pool $\mathcal C_b$. This reduces the lexical search space while preserving candidates that exhibit strong LoRA-subspace alignment. Pool construction details are provided in Appendix~\ref{app:atomic-pool-details}.
\\
\textbf{Prompt-level alignment.}
LoRA-subspace alignment is context dependent, so Atomics that score highly in isolation may no longer receive high scores when combined into a prompt. SGES therefore re-evaluates each Atomic in the context of the complete prompt. Each present site group contributes three Atomics:
\begin{equation}
p
=
\left[
a_b^i
\right]_{b\in\mathcal G_{\mathcal A},\,i=1,2,3},
\qquad
a_b^i\in\mathcal C_b.
\label{eq:typed-prompt}
\end{equation}
We instantiate the abstract alignment functional $\mathcal H(p,\mathcal S_{\mathcal A})$ in Equation~\eqref{eq:lora-subspace-alignment} using the Subspace Alignment Score (SAS):
\begin{equation}
\widehat{\operatorname{Align}}_{\mathcal A}(p)
\triangleq
\operatorname{SAS}(p)
=
\frac{1}{3|\mathcal G_{\mathcal A}|}
\sum_{b\in\mathcal G_{\mathcal A}}
\sum_{i=1}^{3}
R_b(a_b^i\mid p).
\label{eq:prompt-level-sas}
\end{equation}

SAS measures prompt-level LoRA-subspace alignment after contextual interaction among its Atomics.
\\
\textbf{Evolutionary prompt search.}
Although the Atomic pools substantially reduce the search space, exhaustive enumeration remains impractical because $\operatorname{SAS}(p)$ depends on the complete prompt context. SGES therefore performs evolutionary search~\cite{miller1995genetic} using $\operatorname{SAS}(p)$ as the fitness function. The resulting high-alignment prompts form the shortlist $\mathcal P_{\mathrm{short}}$, which is passed to PVI for visual identification. The evolutionary
operators and shortlist construction are detailed in Appendix~\ref{app:ea-shortlist}.

\begin{figure}[t]
    \centering
    \includegraphics[width=\linewidth]{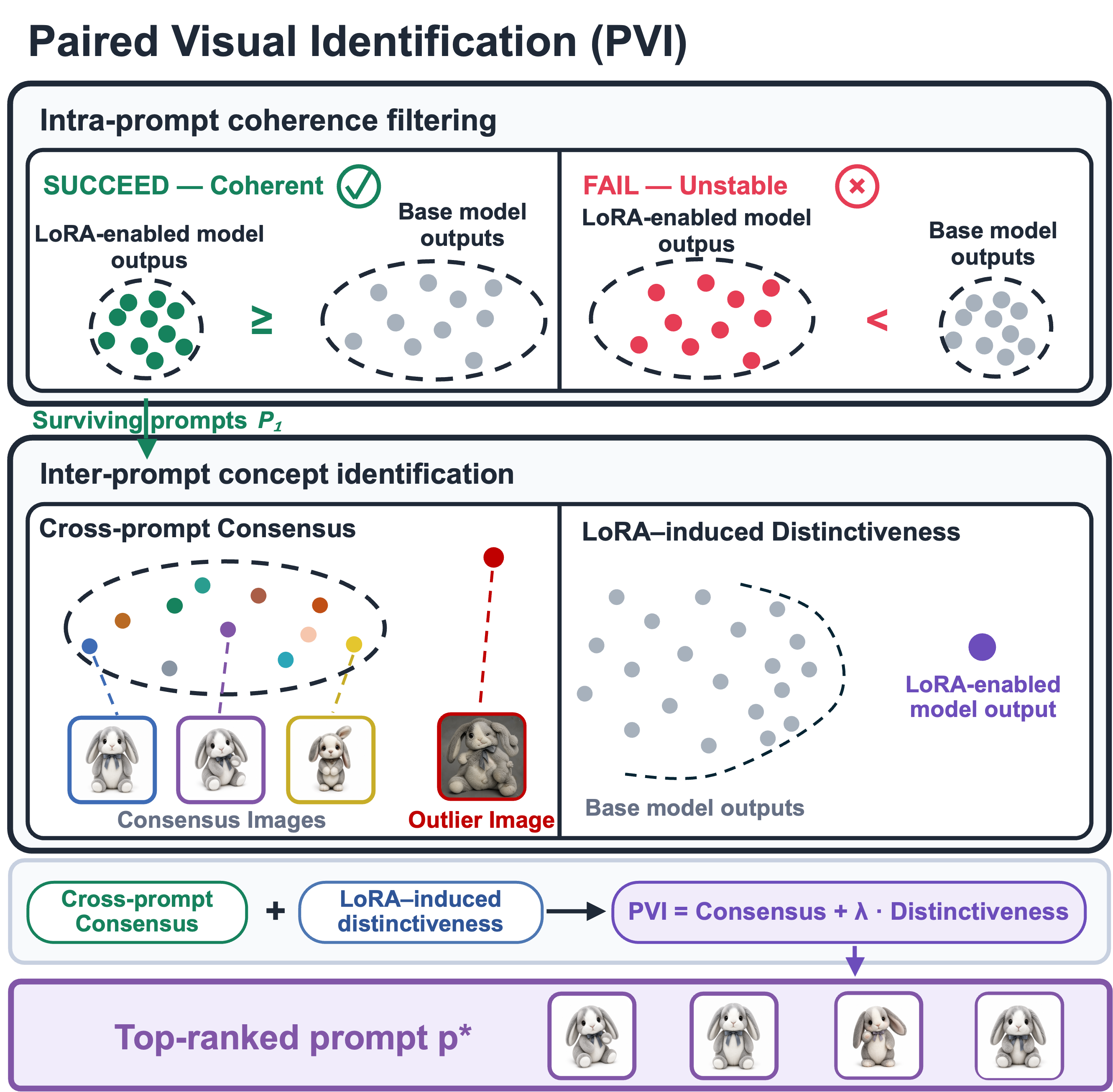}
    \caption{Two-stage PVI workflow. Intra-prompt coherence
    filtering first forms $\mathcal P_1$ by retaining prompts whose LoRA-enabled
    outputs are at least as coherent as the corresponding outputs from the
    compatible base model under matched seeds. Inter-prompt concept
    identification then ranks the survivors using two complementary signals:
    cross-prompt consensus down-ranks isolated candidates whose images are
    visually anomalous relative to the recurring cross-prompt pattern, while
    LoRA-induced distinctiveness favors prompts whose LoRA-enabled outputs are
    separated from the base-model outputs, providing evidence of
    LoRA-specific behavior. Their combined PVI score returns the top-ranked
    complete prompt $p^\star$ together with its LoRA-enabled generated images.}
    \label{fig:pvi-workflow}
\end{figure}

\subsection{Paired Visual Identification}
\label{sec:pvi}
SGES reduces the full prompt space to candidates with strong LoRA-subspace alignment, but alignment alone does not guarantee that their generated outputs reveal a coherent LoRA-specific concept.
PVI therefore evaluates the two output-side properties in Equation~\eqref{eq:dominant-hidden-concept}: output coherence and LoRA-induced distinctiveness. It first filters unstable LoRA responses, then uses cross-prompt consensus to suppress visual outliers, and finally combines consensus with LoRA-induced distinctiveness to identify the dominant carrier. Figure~\ref*{fig:pvi-workflow} mirrors these two stages.
\\
\textbf{Paired output samples.} For each prompt $p\in\mathcal P_{\mathrm{short}}$, PVI evaluates $\mathcal B_{\mathcal A}$ and $\mathcal B$ using the same $R$ random seeds $\{z_r\}_{r=1}^{R}$. We denote the resulting realizations of $\mathbf Y_{\mathcal A}(p)$ and $\mathbf Y_{\mathcal B}(p)$ from Equations~\eqref{eq:lora-conditional-semantic-output} and~\eqref{eq:base-conditional-semantic-output} by $\mathbf y_{\mathcal A}^{r}(p)$ and $\mathbf y_{\mathcal B}^{r}(p)$, respectively. Using matched seeds controls diffusion randomness when comparing the LoRA-enabled and base models~\cite{glasserman1992some}.
\\
\textbf{Intra-prompt coherence filtering.}
We instantiate the coherence functional $\mathcal C(\cdot)$ in Equation~\eqref{eq:output-coherence} using the average pairwise cosine similarity across random generations. For $X\in\{\mathcal A,\mathcal B\}$, we define
\begin{equation}
\widehat{\operatorname{Coh}}_{X}(p)
\triangleq
\frac{2}{R(R-1)}
\sum_{1\leq r<r'\leq R}
\left(
\mathbf y_X^{r}(p)
\right)^\top
\mathbf y_X^{r'}(p).
\label{eq:pvi-coherence}
\end{equation}

A fixed coherence threshold is unsuitable because different concepts naturally exhibit different output variation. We therefore use the compatible base model under the same prompt as an adaptive reference. 
Since the adapter is trained to reinforce its encoded concept, a revealing prompt should yield LoRA-enabled outputs at least as coherent as the corresponding outputs from the compatible base model. We thus retain
\begin{equation}
\mathcal P_1
=
\left\{
p\in\mathcal P_{\mathrm{short}}
\;\middle|\;
\widehat{\operatorname{Coh}}_{\mathcal A}(p)
\geq
\widehat{\operatorname{Coh}}_{\mathcal B}(p)
\right\}.
\label{eq:pvi-level1}
\end{equation}
\\
\textbf{Inter-prompt concept identification.}
After intra-prompt filtering, let $\mathcal P_1=\{p_1,\ldots,p_M\}$ denote the surviving candidates. Although their LoRA-enabled outputs are stable, some candidates may still be visual outliers.
Such outliers can appear highly distinctive from the base model without representing the dominant concept.
PVI therefore combines cross-prompt consensus with LoRA-induced distinctiveness for concept identification.

For each candidate $p_i$, we define its cross-prompt consensus as
\begin{equation}
\operatorname{Con}_{\mathcal A}(p_i)
=
\frac{1}{M-1}
\sum_{j\neq i}
\frac{1}{R}
\sum_{r=1}^{R}
\left(
\mathbf y_{\mathcal A}^{r}(p_i)
\right)^\top
\mathbf y_{\mathcal A}^{r}(p_j).
\label{eq:pvi-lora-consensus}
\end{equation}
A larger $\operatorname{Con}_{\mathcal A}(p_i)$ indicates stronger
agreement with the other surviving candidates and thus a lower
likelihood of being a visual outlier. Outputs are compared under
matched random seeds to reduce variation from diffusion randomness.

We next instantiate the LoRA-induced distinctiveness in Equation~\eqref{eq:lora-induced-distinctiveness} using the average cosine dissimilarity between the LoRA-enabled outputs of $p_i$ and the base-model outputs across the surviving prompts:
\begin{equation}
\widehat{\operatorname{Dist}}_{\mathcal A,\mathcal B}(p_i)
\triangleq
\frac{1}{M}
\sum_{j=1}^{M}
\frac{1}{R}
\sum_{r=1}^{R}
\left[
1-
\left(
\mathbf y_{\mathcal A}^{r}(p_i)
\right)^\top
\mathbf y_{\mathcal B}^{r}(p_j)
\right].
\label{eq:pvi-distinctiveness}
\end{equation}
A larger $\widehat{\operatorname{Dist}}_{\mathcal A,\mathcal B}(p_i)$ indicates stronger separation from the base-model outputs.

PVI combines cross-prompt consensus with the empirical distinctiveness score above:
\begin{equation}
\operatorname{PVI}(p_i)
=
\operatorname{Con}_{\mathcal A}(p_i)
+
\lambda
\widehat{\operatorname{Dist}}_{\mathcal A,\mathcal B}(p_i),
\label{eq:pvi-score}
\end{equation}
where $\lambda\geq0$ balances the two terms. The highest-scoring candidate is returned as the dominant carrier prompt $p^\star$.

\section{Evaluation}
\label{sec:evaluation}

\subsection{Evaluation Setup}
\label{sec:evaluation-setup}
\textbf{Benchmark and models.}
We evaluate \sys on LoRA adapters from the SD1.5 and SDXL model
families~\cite{rombach2022high,podell2024sdxl,stablediffusion2026v15modelcard,stabilityai2026sdxlmodelcard}. The benchmark spans real identities, synthetic identities,
anime characters, and artistic styles, and combines public adapters~\cite{civitai2026models,huggingface2026modelhub}
with identity adapters trained under controlled settings. Each adapter
is paired with a compatible base model from the same model family.
Appendix~\ref{app:benchmark-details} gives the benchmark composition;
Appendices~\ref{app:controlled-training} and~\ref{app:identity-data}
give the local training and data-construction procedures. We keep these
cohort details in the appendix and state below whenever a result uses a
restricted subset.
\\
\textbf{Baselines.}
Hidden-concept discovery requires identifying an unknown concept without a predefined concept, and few existing methods directly address this setting. To enable broad comparison, we evaluate seven representative methods from closely related tasks. These methods can be broadly grouped into two categories:
\vspace{-6pt}
\begin{itemize}
\setlength{\topsep}{0pt}
\setlength{\itemsep}{1pt}
\setlength{\parsep}{0pt}
\setlength{\parskip}{0pt}
\setlength{\partopsep}{0pt}

\item \emph{Concept-agnostic difference discovery.}
DriftScope~\cite{laria2026driftscope}, CompCon~\cite{dunlap2025discovering}, VisDiff~\cite{dunlap2024describing}, and DPO-Diff-EPO~\cite{wang2024discrete} share the goal of exposing differences between model behaviors without assuming a predefined concept. We adapt them to compare the LoRA-enabled model with its compatible base model and interpret the discovered LoRA-specific differences as candidate hidden concepts.

\item \emph{Concept-specified verification.}
PEZ~\cite{wen2023hard} recovers prompts from example images of a specified concept, which we use to query the LoRA-enabled model. PAIA~\cite{yuan2025lurks} instead directly verifies whether a specified concept is encoded by comparing the denoising behavior of the LoRA-enabled and base models. All three methods require the concept to be specified beforehand, so their coverage is limited to the candidate concepts examined.
\end{itemize}
\vspace{-6pt}
Detailed baseline adaptations are provided in Appendix~\ref{app:baseline-details}.
\textbf{Evaluation metric.}
Equation~\eqref{eq:target-concept-discrepancy} defines the discrepancy
between the outputs induced by a discovered carrier prompt and the
semantic direction of a target concept. In evaluation, let
$\mathcal Y_{p^\star}=\{\mathbf y_i\}_{i=1}^{n}$ and
$\mathcal R_c=\{\mathbf r_j\}_{j=1}^{m}$ denote finite samples from
$Q_{\mathcal A}(\cdot\mid p^\star)$ and $Q_c^{\mathrm{ref}}$,
respectively. We estimate the reference concept direction by
\begin{equation}
\widehat{\boldsymbol{\mu}}_c^{\mathrm{ref}}
=
\frac{1}{m}
\sum_{j=1}^{m}
\mathbf r_j,
\qquad
\overline{\boldsymbol{\mu}}_c^{\mathrm{ref}}
=
\frac{
\widehat{\boldsymbol{\mu}}_c^{\mathrm{ref}}
}{
\left\|
\widehat{\boldsymbol{\mu}}_c^{\mathrm{ref}}
\right\|_2
}.
\label{eq:empirical-target-concept-direction}
\end{equation}
The empirical target-concept discrepancy is then
\begin{equation}
\widehat{\Delta}_{\mathcal A}(p^\star;c)
\triangleq
1-
\frac{1}{n}
\sum_{i=1}^{n}
\mathbf y_i^\top
\overline{\boldsymbol{\mu}}_c^{\mathrm{ref}}.
\label{eq:empirical-target-concept-discrepancy}
\end{equation}
Since the semantic representations are $\ell_2$ normalized,
$\widehat{\Delta}_{\mathcal A}(p^\star;c)$ is the average cosine
distance between the generated outputs and the estimated semantic
direction of concept $c$. A smaller discrepancy indicates that the
discovered carrier more faithfully reveals the target concept.
\\
\textbf{Implementation details.}
SGES retains the top $k=2$ right singular directions at each selected
LoRA site. Each present site group retains 50 Atomic candidates,
including 30 token candidates and 20 phrase candidates. Evolutionary
search uses a population of 48 for 100 generations and is repeated over
three independent runs. PVI uses $R=5$ paired generation seeds and sets
$\lambda=0.5$. We use \texttt{openai/clip-vit-large-patch14}~\cite{radford2021learning,openai2026clipvitl14} for PVI
and final evaluation. For each evaluated prompt, we generate $n=10$
images and separately generate $m=10$ held-out references using
disjoint random seeds; references are never available during discovery.
All image generation follows the compatible configuration of each
benchmark case.
Experiments run on NVIDIA A40 GPUs. The precise
text-conditioning sites, composed-word procedure, Atomic-pool
construction, and evolutionary shortlist are documented in
Appendices~\ref{app:sges-sites}--\ref{app:ea-shortlist}.

\raggedbottom
\subsection{Main Results}
\label{sec:main-results}
\flushbottom

\subsubsection{Hidden-Concept Discovery}
\label{sec:hidden-concept-discovery}

\begin{table}[t]
\centering
\caption{Hidden-concept discovery on the common faithful SD1.5 cohort.
Concept Input indicates access to target examples. Entries report mean
target-concept discrepancy within each category; Mean is
adapter-weighted across adapters. Lower is better.}
\label{tab:main-comparison}
\small
\setlength{\tabcolsep}{1.5pt}
\begin{tabular}{@{}lccccc@{}}
\toprule
Method &
\shortstack{Concept\\Input} &
Real ID $\downarrow$ &
\shortstack{Synthetic\\ID $\downarrow$} &
Anime $\downarrow$ &
Mean $\downarrow$ \\
\midrule
DriftScope
& $\times$
& 0.451
& 0.437
& 0.416
& 0.435 \\
CompCon
& $\times$
& 0.543
& 0.462
& 0.404
& 0.475 \\
VisDiff
& $\times$
& 0.396
& 0.333
& 0.393
& 0.383 \\
DPO-Diff-EPO
& $\times$
& 0.552
& 0.449
& 0.490
& 0.509 \\
PEZ
& $\checkmark$
& \underline{0.171}
& \underline{0.196}
& \underline{0.190}
& \underline{0.183} \\
\textbf{Ours}
& $\times$
& \textbf{0.078}
& \textbf{0.178}
& \textbf{0.123}
& \textbf{0.114} \\
\bottomrule
\end{tabular}
\end{table}

Table~\ref{tab:main-comparison} compares the semantic fidelity of the
returned prompts using the target-concept discrepancy in
Equation~\eqref{eq:target-concept-discrepancy}. DriftScope, CompCon,
VisDiff, DPO-Diff-EPO, and \sys receive no predefined concept during
search. PEZ is instead supplied with target-concept images, providing a
reference-assisted comparison.
\\
\textbf{Results.}
\sys achieves the lowest discrepancy for every category, with an
adapter-weighted mean of 0.114. VisDiff is the
strongest comparison method under the same no-concept-input setting,
with a mean of 0.383; \sys reduces this discrepancy by 70.2\%. PEZ
benefits substantially from its target images and reaches 0.183, yet
\sys further reduces its discrepancy by 37.7\% without receiving a
concept name, trigger, or target example.

The category-level results follow the same ordering. For real
identities, \sys reaches 0.078, compared with 0.396 for VisDiff and
0.171 for PEZ. For synthetic identities, it reaches 0.178, improving
on 0.333 for VisDiff and 0.196 for PEZ. For anime characters, it reaches
0.123, compared with 0.393 and 0.190, respectively. Thus, the overall
gain is supported by all three categories rather than being driven by
one concept type.
\\
\textbf{Analysis.}
The separation from the concept-agnostic baselines highlights the
difference between exposing an arbitrary LoRA--base change and
identifying the adapter's dominant hidden concept. A LoRA can alter
several concepts or visual attributes, so a prompt that maximizes a
behavioral difference need not reveal the encoded target concept.
\sys adds two complementary sources of selectivity: SGES prioritizes
prompts supported by the leading input subspaces of the LoRA updates,
and PVI retains coherent responses that recur across high-alignment
prompts and remain distinctive from the compatible base model. The
component results in Section~\ref{sec:ablation-components} directly
isolate these contributions.

PEZ provides an informative second comparison. Its improvement over
the concept-agnostic baselines confirms that target images offer strong
guidance for prompt recovery. \sys nevertheless achieves a lower
discrepancy in every category without this guidance. This result is
consistent with the paper's central premise: the adapter parameters
themselves contain useful evidence for localizing the dominant concept,
and output-side identification can turn that evidence into a prompt
whose generations match the target.
The comparison therefore tests the deployment setting directly:
whether an unknown target can be localized from a submitted adapter,
rather than whether a supplied target can be reproduced.

\subsubsection{Concept-Specificity Verification}
\label{sec:concept-verification}

\begin{table}[t]
\centering
\caption{Concept-specificity verification with one target and one fixed
non-target query per adapter. PEZ and PAIA receive the queried concept,
whereas \sys uses its fixed discovered prompt. Higher is better.}
\label{tab:concept-verification}
\small
\setlength{\tabcolsep}{4pt}
\begin{tabular}{@{}lcccc@{}}
\toprule
Method &
Accuracy $\uparrow$ &
Precision $\uparrow$ &
Recall $\uparrow$ &
F1 $\uparrow$ \\
\midrule
PEZ
& 0.941
& 0.959
& 0.922
& 0.940 \\
PAIA
& 0.951
& 0.911
& \textbf{1.000}
& 0.953 \\
\textbf{Ours}
& \textbf{0.980}
& \textbf{0.980}
& 0.980
& \textbf{0.980} \\
\bottomrule
\end{tabular}
\end{table}

The discrepancy experiment above measures how closely a returned
prompt reveals the target. We next test the complementary property of
concept specificity: whether the same evidence distinguishes the
target from a non-target concept instead of merely expressing a broad
category or incidental attribute. For each adapter, we form one
positive query using its ground-truth concept and one negative query
using a deterministically selected non-target concept. This produces a
balanced evaluation with identical queries for all methods.

PEZ and PAIA receive the queried concept before making each decision.
In contrast, \sys first discovers $p^\star$ without a concept query and
then uses that same fixed prompt for both its target and non-target
decision. For PEZ and \sys, we apply the pre-specified threshold
$\tau=0.3$ to the target-concept discrepancy:
\begin{equation}
\widehat y(\mathcal A_i,c)
=
\mathbbm 1
\left[
\widehat{\Delta}_{\mathcal A_i}(p;c)
\leq \tau
\right],
\qquad \tau=0.3.
\label{eq:concept-verification-decision}
\end{equation}
Here $p=p^\star$ for \sys and is the prompt recovered from the supplied
query images for PEZ. PAIA uses its native binary decision rule.
\\
\textbf{Results.}
Table~\ref{tab:concept-verification} shows that \sys achieves the
highest accuracy, precision, and F1, all at 0.980, while also reaching
0.980 recall. PAIA attains perfect recall but has lower precision of
0.911, indicating that its operating point favors positive decisions.
PEZ has higher precision of 0.959 than recall of 0.922, showing the
opposite imbalance. In comparison, \sys obtains equal precision and
recall while improving overall accuracy and F1. The advantage therefore
comes from treating target and non-target decisions more evenly rather
than optimizing only one side of the binary task.
\\
\textbf{Analysis.}
These results complement the continuous fidelity measurement in
Table~\ref{tab:main-comparison}. A low discovery discrepancy shows that
the returned prompt reveals the intended concept, while high
specificity additionally requires rejecting a controlled non-target.
The 0.980 precision and recall show that the fixed $p^\star$ supports
both properties. In particular, the result is obtained without
query-conditioned search: \sys discovers the prompt once and then
evaluates it against both concepts.

PEZ and PAIA begin from the concept being tested, whereas \sys must
first determine what concept the adapter predominantly encodes. Its
higher accuracy and F1 therefore show that open-ended discovery can
produce evidence useful for a subsequent concept-specific decision.
Together, Tables~\ref{tab:main-comparison} and
\ref{tab:concept-verification} establish two distinct properties of the
returned prompt: semantic fidelity to the hidden target and specificity
against a non-target concept.

\subsubsection{Composition of the Top-Ranked Prompt}
\label{sec:carrier-composition}

\begin{figure*}[t]
    \centering
    \includegraphics[width=\linewidth]{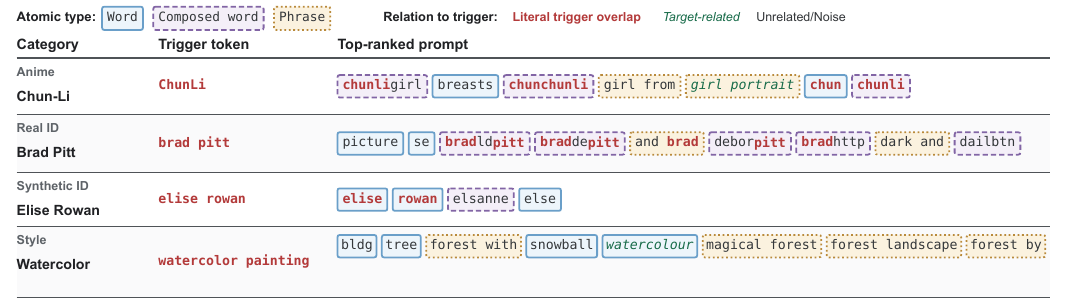}
    \caption{Composition of representative top-ranked prompts. The first
    column reports each LoRA category and concept, the second gives its
    training trigger revealed only after discovery, and the third
    decomposes $p^\star$ returned by \sys. Solid blue, dashed purple, and
    dotted ochre borders denote Word, Composed-word, and Phrase Atomics.
    In $p^\star$, red text marks literal trigger overlap, italic green text
    marks target-related descriptions, and black text marks
    unrelated/noise. All annotations are added post hoc and are never
    used by discovery.}
    \label{fig:carrier-composition}
\end{figure*}

Figure~\ref{fig:carrier-composition} shows how the content of the
top-ranked prompt relates to the training trigger and encoded concept.
For Chun-Li, the complete \texttt{chunli} string occurs inside several
Composed-word Atomics; for Brad Pitt, \texttt{brad} and \texttt{pitt}
recur inside different Composed-word Atomics. These examples demonstrate
the value of BPE-aware composition for names represented by multiple
lexical units. For Elise Rowan, both trigger words are recovered as
separate Word Atomics. The Watercolor example follows a complementary
semantic route, combining \texttt{watercolour} with target-related
descriptions.

Because SGES selects Atomics by alignment with the dominant LoRA input
subspaces, literal fragments and semantic descriptions can jointly
activate the encoded concept. Exact trigger recovery is not the
discovery objective: a carrier prompt succeeds when its LoRA-enabled
outputs reveal the encoded concept. Semantically related or partially
overlapping Atomics can therefore be sufficient even when $p^\star$
does not reconstruct the training trigger.

\subsubsection{Visual Discovery Results}
\label{sec:qualitative-discovery}

\begin{figure*}[t]
    \centering
    \includegraphics[width=\linewidth]{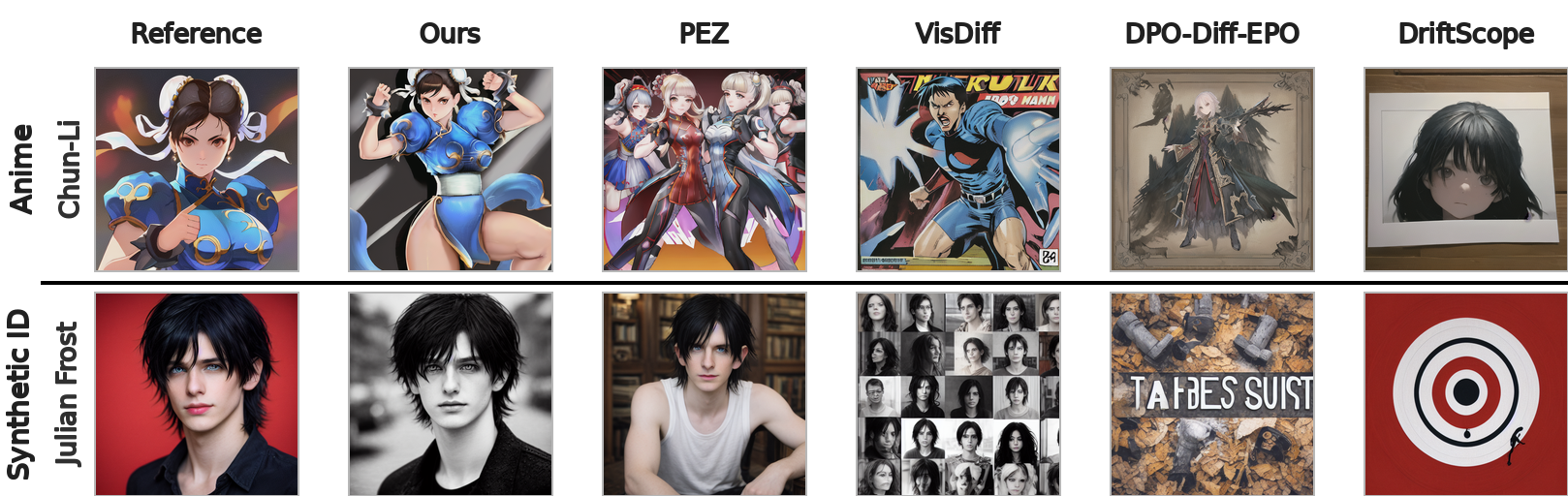}
    \caption{Visual comparison of hidden-concept discovery for an anime
    character and a synthetic identity. Reference images are generated
    by the LoRA-enabled model under verified trigger prompts; the
    remaining columns show LoRA-enabled generations under each method's
    returned prompt. PEZ receives target-concept images, whereas \sys and
    the other displayed methods receive no target-concept input.}
    \label{fig:qualitative-baselines}
\end{figure*}

Figure~\ref{fig:qualitative-baselines} makes the quantitative result
visually concrete. For both Chun-Li and Julian Frost, generations from
\sys's top-ranked prompts preserve the defining identity visible in
the references. PEZ is the closest comparison method and is additionally
given target-concept images; in the anime example, however, it recovers
the anime domain rather than the same character. In the examples shown,
the remaining methods return different identities, characters, or
incidental visual patterns. In contrast, \sys reveals both hidden
identities without any target-concept information, showing that the
parameter-guided search recovers the dominant LoRA concept rather than
only a conspicuous output difference.

\subsubsection{Discovery Across Backbones and Concept Types}
\label{sec:cross-family-generalization}

\begin{figure*}[t]
    \centering
    \begin{minipage}[t]{0.49\textwidth}
        \centering
        \includegraphics[width=\linewidth]{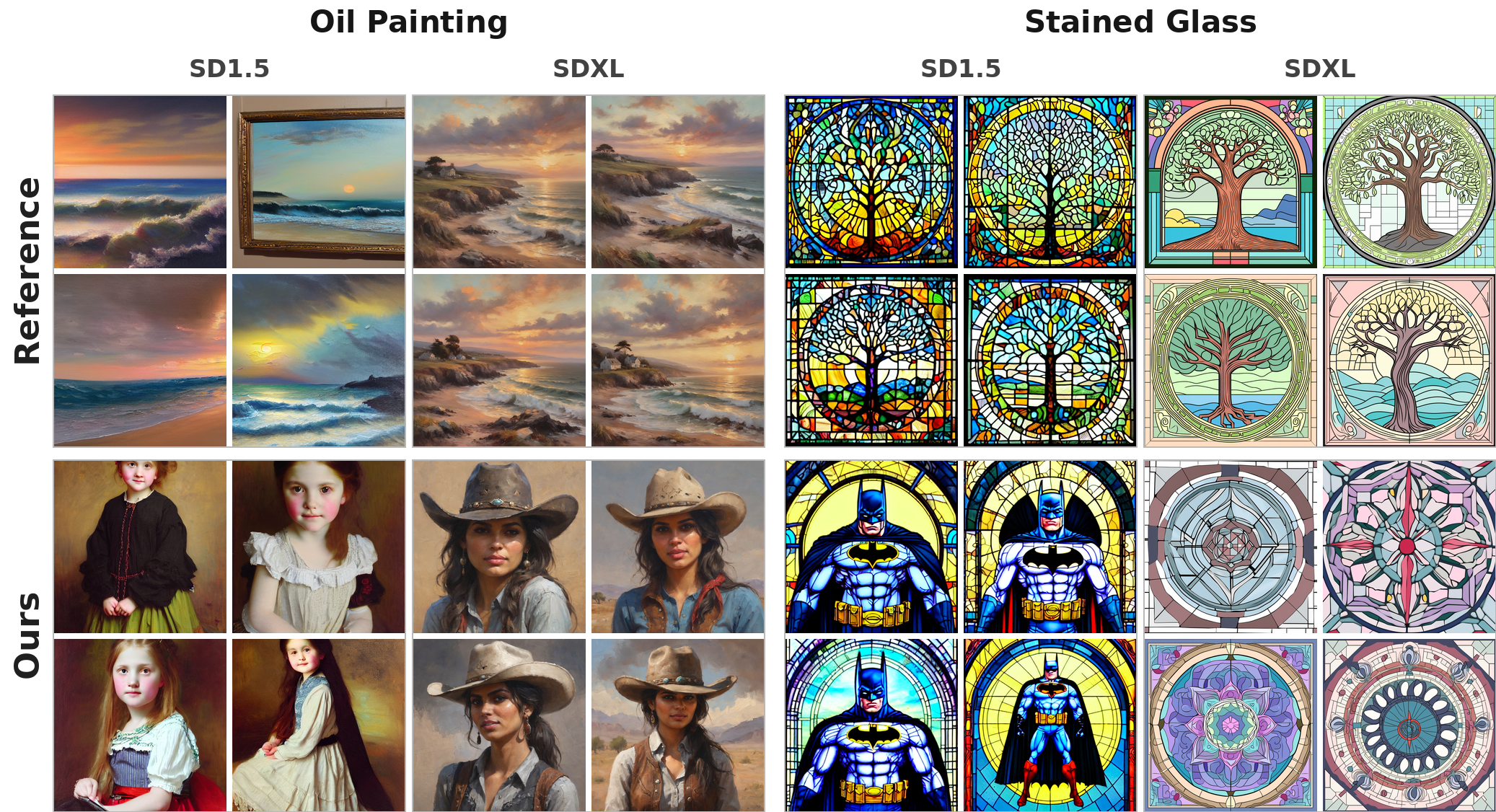}
        \vspace{2pt}
        \small (a) Qualitative style discovery
    \end{minipage}
    \hfill
    \begin{minipage}[t]{0.49\textwidth}
        \centering
        \includegraphics[width=\linewidth]{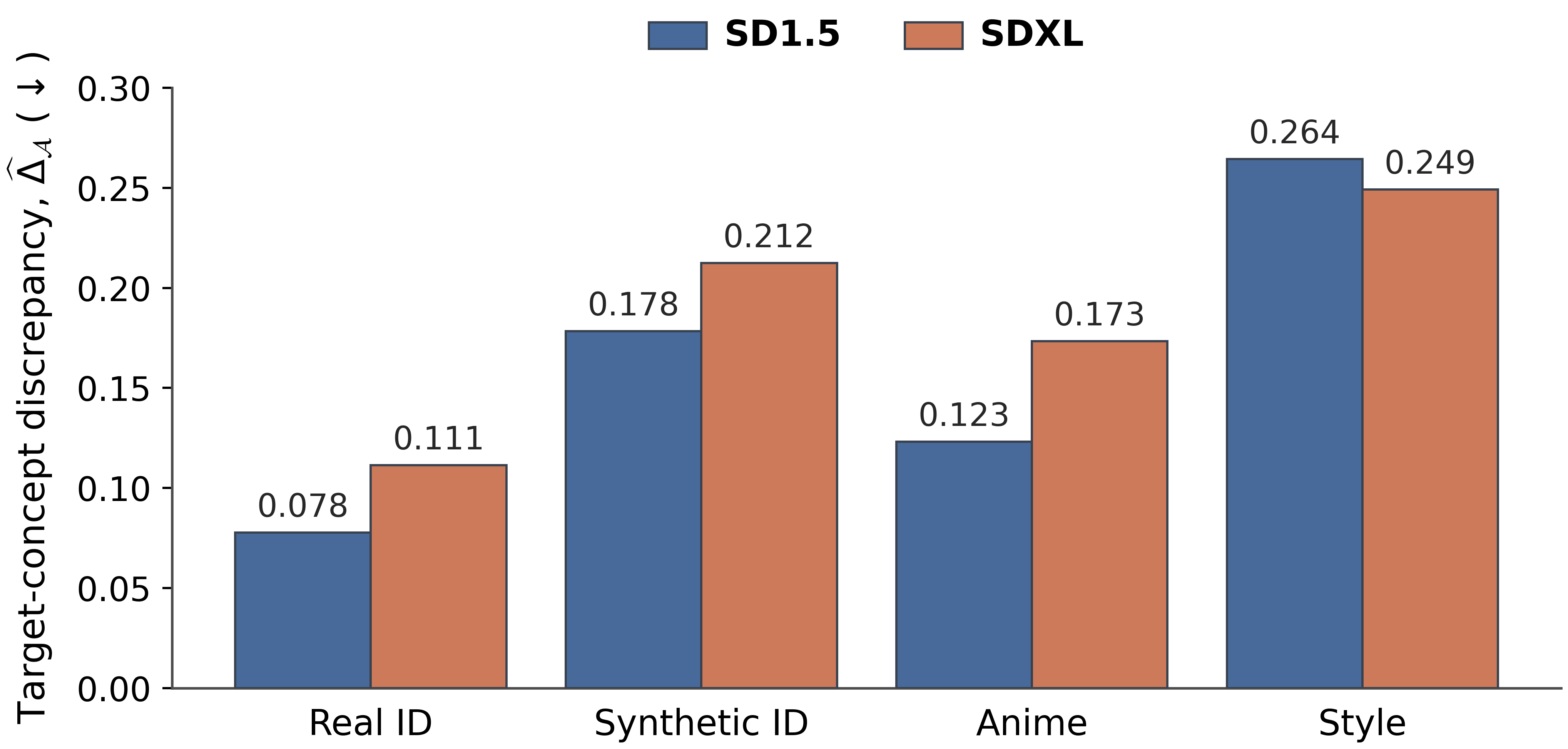}
        \vspace{2pt}
        \small (b) Discrepancy by concept type and backbone
    \end{minipage}
    \caption{Discovery across backbones and concept types.
    (a) Trigger-prompt generations (top) and generations from \sys's
    top-ranked prompts (bottom) for two styles on SD1.5 and SDXL.
    (b) Mean target-concept discrepancy by concept type and backbone;
    lower is better.}
    \label{fig:style-evaluation}
\end{figure*}

Figure~\ref{fig:style-evaluation}(b) shows that the same discovery
procedure produces concept-aligned prompts on both SD1.5 and SDXL, with
all concept-type means below 0.3. For real identities, synthetic
identities, and anime characters, discrepancies are 0.078, 0.178, and
0.123 on SD1.5 and 0.111, 0.212, and 0.173 on SDXL. Style discovery
reaches 0.264 on SD1.5 and 0.249 on SDXL, extending the method to
concepts defined by shared visual attributes rather than one identity.

Figure~\ref{fig:style-evaluation}(a) makes this result visible. The
oil-painting prompts reproduce painterly texture and color treatment
across different subjects, while the stained-glass prompts reproduce
segmented geometry, dark leading, and luminous color across different
scenes. The returned prompts therefore recover the shared high-level
style instead of copying one reference subject.

\subsection{Ablation Study}
\label{sec:ablation}

We isolate the contribution of the two pipeline stages, the LoRA input
directions used by SGES, contextual prompt search, and site-specific
Atomic pools. Appendix~\ref{app:ablation-k} additionally evaluates the
retained subspace dimension.

\subsubsection{Impact of the Core Components}
\label{sec:ablation-components}

\begin{table}[t]
\centering
\caption{Contributions of SGES, PVI, composed-word discovery, and the
three PVI operations. Each row removes one component and reruns
discovery; lower target-concept discrepancy is better.}
\label{tab:component-ablation}
\small
\setlength{\tabcolsep}{2.5pt}
\begin{tabular}{@{}lcccc@{}}
\toprule
Variant &
\shortstack{Real ID $\downarrow$} &
\shortstack{Synthetic ID $\downarrow$} &
Anime $\downarrow$ &
Mean $\downarrow$ \\
\midrule
w/o SGES
& 0.472 & 0.313 & 0.344 & 0.393 \\
w/o PVI
& 0.096 & 0.192 & 0.138 & 0.130 \\
w/o CW
& 0.092 & 0.180 & 0.126 & 0.122 \\
w/o Coh.
& 0.087 & 0.187 & 0.131 & 0.123 \\
w/o Cons.
& 0.164 & 0.238 & 0.242 & 0.208 \\
w/o Dist.
& 0.083 & 0.179 & 0.128 & 0.119 \\
\textbf{Full}
& \textbf{0.078}
& \textbf{0.178}
& \textbf{0.123}
& \textbf{0.114} \\
\bottomrule
\end{tabular}
\end{table}

CW denotes composed-word discovery; Coh., Cons., and Dist.\ denote
intra-prompt coherence filtering, cross-prompt consensus, and
LoRA-induced distinctiveness. Composed-word discovery first enriches
the lexical search space: removing it raises the mean from 0.114 to
0.122, with the largest change for real identities (0.078 to 0.092).
This agrees with Figure~\ref{fig:carrier-composition}, where
multi-unit names are recovered through composed Atomics.

SGES then converts this richer vocabulary into parameter-supported
prompt candidates. Removing SGES and sampling from the raw lexical
source raises the mean to 0.393, the largest change in
Table~\ref{tab:component-ablation}. PVI acts on the resulting SGES
shortlist; removing it raises the mean to 0.130, showing that
output-side identification adds a consistent gain after
parameter-guided search.

The PVI ablations follow its processing order. Intra-prompt coherence
filtering removes unstable candidates and improves the mean from 0.123
to 0.114. Cross-prompt consensus provides the largest identification
gain: without it, the mean rises to 0.208 and anime discrepancy rises
from 0.123 to 0.242, confirming the value of down-ranking visual
outliers that do not recur across other high-alignment prompts.
LoRA-induced distinctiveness then favors outputs introduced by the
adapter, improving the mean from 0.119 to 0.114. Together, the
components form a connected progression from BPE-aware candidates, to
parameter-guided localization, to visual identification.

\subsubsection{Impact of the LoRA Subspace Selection}
\label{sec:ablation-directions}

\begin{table}[t]
\centering
\caption{Effect of LoRA input-direction selection at $k=2$. Random
uses a dimension-matched orthonormal subspace; lower target-concept
discrepancy is better.}
\label{tab:ablation-directions}
\small
\setlength{\tabcolsep}{3.5pt}
\begin{tabular}{@{}lcccc@{}}
\toprule
Direction &
Real ID $\downarrow$ &
Synthetic ID $\downarrow$ &
Anime $\downarrow$ &
Mean $\downarrow$ \\
\midrule
Random
& 0.396 & 0.258 & 0.301 & 0.334 \\
Trailing
& 0.485 & 0.343 & 0.327 & 0.398 \\
\textbf{Leading}
& \textbf{0.078}
& \textbf{0.178}
& \textbf{0.123}
& \textbf{0.114} \\
\bottomrule
\end{tabular}
\end{table}

This experiment tests the parameter-space signal at the center of
SGES. We fix $k=2$ and replace the leading right-singular directions
with either the final two directions in the effective update subspace
or a dimension-matched random orthonormal subspace.

Table~\ref{tab:ablation-directions} gives the same strict ordering in
every category. Leading directions achieve a mean discrepancy of 0.114,
compared with 0.334 for random directions and 0.398 for trailing
directions. Because the random control has the same dimension, low
dimensionality alone does not explain the gain. Singular-value ordering
concentrates the LoRA evidence that SGES uses to localize
hidden-concept prompts. The dimension sensitivity in
Appendix~\ref{app:ablation-k} further gives means of 0.128, 0.114, and
0.127 for $k=1,2,4$, respectively; $k=2$ is best, and the tested range
varies by at most 0.014.

\subsubsection{Impact of the Search Strategy}
\label{sec:ablation-search}

\begin{table}[t]
\centering
\caption{Prompt-search strategies over identical group-specific Atomic
pools with fixed shortlist and PVI budgets. Lower target-concept
discrepancy is better.}
\label{tab:ablation-search}
\small
\setlength{\tabcolsep}{2pt}
\begin{tabular}{@{}lcccc@{}}
\toprule
Strategy &
Real ID $\downarrow$ &
Synthetic ID $\downarrow$ &
Anime $\downarrow$ &
Mean $\downarrow$ \\
\midrule
Greedy
& 0.148 & 0.217 & 0.180 & 0.173 \\
Random
& 0.107 & 0.191 & 0.138 & 0.135 \\
\textbf{Evolutionary}
& \textbf{0.078}
& \textbf{0.178}
& \textbf{0.123}
& \textbf{0.114} \\
\bottomrule
\end{tabular}
\end{table}

This experiment holds the group-specific Atomic pools, shortlist
budget, and PVI stage fixed and changes only how complete prompts are
selected. Greedy combines the strongest standalone Atomics, Random
samples from the already constructed Atomic pools, and Evolutionary
re-encodes each complete prompt and optimizes its contextual SAS. Thus,
Random here differs from \emph{w/o SGES} in
Table~\ref{tab:component-ablation}: it retains SGES's
parameter-enriched Atomic pools rather than sampling from the raw
lexical source.

Table~\ref{tab:ablation-search} shows the same ordering in every
category. Random reaches 0.135, demonstrating that the group-specific
Atomic pools already contain useful signal. Greedy reaches 0.173
because Atomics that align well alone need not remain optimal after
contextual encoding. Evolutionary search reaches 0.114 by optimizing
the complete prompt in context, improving on both fixed alternatives.

\subsubsection{Impact of the Site Grouping}
\label{sec:ablation-buckets}

\begin{table}[t]
\centering
\caption{Contribution of text-encoder (TE), U-Net up-path (UP), and
U-Net down-path (DOWN) site groups. One Bucket merges candidates from
all retained sites; lower target-concept discrepancy is better.}
\label{tab:ablation-buckets}
\small
\setlength{\tabcolsep}{3pt}
\begin{tabular}{@{}lcccc@{}}
\toprule
Variant &
Real ID $\downarrow$ &
Synthetic ID $\downarrow$ &
Anime $\downarrow$ &
Mean $\downarrow$ \\
\midrule
One Bucket
& 0.097 & 0.183 & 0.156 & 0.136 \\
w/o TE
& 0.091 & 0.180 & 0.150 & 0.131 \\
w/o UP
& 0.136 & 0.250 & 0.203 & 0.183 \\
w/o DOWN
& 0.112 & 0.271 & 0.133 & 0.151 \\
\textbf{Full}
& \textbf{0.078}
& \textbf{0.178}
& \textbf{0.123}
& \textbf{0.114} \\
\bottomrule
\end{tabular}
\end{table}

Table~\ref{tab:ablation-buckets} separates access to LoRA sites from
the organization of their evidence. One Bucket retains all selected
sites but merges their candidates, increasing mean discrepancy from
0.114 to 0.136. Separate pools therefore add value even when no site is
removed: they preserve complementary evidence and prevent one pathway
from dominating prompt construction.

The removal results show how those contributions combine. Excluding UP
has the largest overall effect, increasing the mean to 0.183. Excluding
DOWN is especially costly for synthetic identities, increasing their
discrepancy from 0.178 to 0.271, while excluding TE also degrades every
category. These distinct category profiles explain why the full method
constructs prompts from all available site groups.

\subsection{Robustness to Concealed Triggers and Adaptive Training}
\label{sec:robustness}

We evaluate two increasingly targeted challenges: an opaque training
trigger that removes semantic disclosure, and adaptive training that
disperses the dominant LoRA input subspaces used by SGES.

\subsubsection{Discovery Under Opaque Triggers}
\label{sec:unfaithful-disclosure}

\begin{table}[t]
\centering
\caption{Discovery under matched human-readable and opaque-trigger
training. Each pair shares its data and optimization settings and
differs only in the trigger; $\Delta_{\mathrm{trig}}$ is opaque minus
human-readable discrepancy. Lower is better.}
\label{tab:trigger-robustness}
\small
\setlength{\tabcolsep}{3.5pt}
\begin{tabular}{@{}llccc@{}}
\toprule
Model & Concept &
\shortstack{Human-\\readable $\downarrow$} &
Opaque $\downarrow$ &
$\Delta_{\mathrm{trig}}$ $\downarrow$ \\
\midrule
\multirow{3}{*}{SD1.5}
& Real ID
& 0.078 & 0.224 & +0.147 \\
& Synthetic ID
& 0.178 & 0.292 & +0.114 \\
& Overall
& 0.112 & 0.247 & +0.136 \\
\midrule
\multirow{3}{*}{SDXL}
& Real ID
& 0.111 & 0.204 & +0.093 \\
& Synthetic ID
& 0.212 & 0.358 & +0.145 \\
& Overall
& 0.146 & 0.257 & +0.111 \\
\midrule
\multicolumn{2}{l}{\textbf{Combined}}
& \textbf{0.129}
& \textbf{0.252}
& \textbf{+0.123} \\
\bottomrule
\end{tabular}
\end{table}

Each controlled identity is represented by a matched human-readable
and opaque-trigger adapter on both backbones. A pair shares training
images and optimization settings and differs only in its caption
trigger. Neither trigger is available during discovery. For each pair,
we define
\begin{equation}
\Delta_{\mathrm{trig}}
=
\widehat{\Delta}_{\mathcal A_{\mathrm{opaque}}}
(p^\star_{\mathrm{opaque}};c)
-
\widehat{\Delta}_{\mathcal A_{\mathrm{readable}}}
(p^\star_{\mathrm{readable}};c).
\label{eq:trigger-gap}
\end{equation}

Table~\ref{tab:trigger-robustness} shows that replacing readable names
with opaque strings increases the combined mean discrepancy from 0.129
to 0.252. Nevertheless, the opaque-trigger means remain 0.247 on SD1.5
and 0.257 on SDXL, and the combined mean remains below the evaluation
threshold of 0.3. Thus, \sys continues to expose substantial concept
evidence even when the training trigger provides no human-readable
semantics. Because neither trigger is supplied during discovery, this
robustness comes from evidence encoded in the adapter rather than
direct access to the trigger text.

\subsubsection{Adaptive Training: Subspace Dispersion}
\label{sec:adaptive-attacks}

We next evaluate a creator who knows that SGES uses the dominant LoRA
input subspaces and adds a fixed spectral-dispersion regularizer during
training. The creator does not know the retained dimension, inspected
sites, or other discovery hyperparameters. The attack counts as a
successful evasion only when the creator's fixed trigger prompt still
reveals the target but the top-ranked prompt returned by \sys does not.
Appendix~\ref{app:adaptive-attacks} gives the threat model, objective,
schedule, and matched controls.

For an attacked adapter, $p_{\mathrm{trig}}$ denotes the fixed
generation prompt containing its human-readable training trigger; it is
available to the creator but unavailable during discovery. \sys instead
returns $p^\star$. \emph{Utility Preserved} is the fraction satisfying
$\widehat{\Delta}_{\mathcal A}(p_{\mathrm{trig}};c)\leq0.3$.
\emph{Conditional Evasion} is the fraction of these utility-preserving
adapters for which
$\widehat{\Delta}_{\mathcal A}(p^\star;c)>0.3$.

\begin{table}[t]
\centering
\caption{Adaptive subspace-dispersion attack. Prompt columns report mean
target-concept discrepancy under attacked adapters. Conditional
evasion is measured only among adapters whose fixed trigger prompt
preserves target generation.}
\label{tab:adaptive-attacks}
\small
\setlength{\tabcolsep}{1.2pt}
\begin{tabular}{@{}lcccc@{}}
\toprule
Attack &
\shortstack{Trigger\\Prompt $\downarrow$} &
\shortstack{Top-ranked\\Prompt $\downarrow$} &
\shortstack{Utility\\Preserved $\uparrow$} &
\shortstack{Conditional\\Evasion $\downarrow$} \\
\midrule
Dispersion & 0.153 & 0.162 & 86.2\% & 4.0\% \\
\bottomrule
\end{tabular}
\end{table}

\textbf{Attack effectiveness.}
The mean spectral-concentration statistic in
Equation~\eqref{eq:module-spectral-concentration} falls from 0.127 to
0.013, an 89.8\% reduction. The attack therefore substantially flattens
the LoRA spectrum and directly weakens the structure SGES is designed
to use.

\textbf{Utility-preserving evasion.}
The attacked adapters retain target-generation utility in 86.2\% of
cases. Their mean trigger-prompt and top-ranked-prompt discrepancies
are 0.153 and 0.162, respectively. Although dispersion raises the
top-ranked-prompt discrepancy from 0.112 on the matched clean cohort to
0.162, conditional evasion among utility-preserving adapters is only
4.0\%. Under this fixed attack, even an 89.8\% reduction in spectral
concentration rarely conceals the target while preserving the
creator's ability to generate it.

\flushbottom

\section{Discussion}
\label{sec:discussion}

\textbf{Why parameter-guided discovery works.}
The experiments show that LoRA parameters provide useful evidence for
open-ended concept discovery even when the concept name, training
trigger, and reference images are unavailable. SGES uses the dominant
LoRA input subspaces to concentrate search on parameter-supported
prompts. PVI then applies intra-prompt coherence filtering,
cross-prompt consensus, and LoRA-induced distinctiveness to select the
prompt whose outputs most consistently reveal the adapter-specific
concept. The ablations connect these roles: removing SGES produces the
largest overall degradation, while cross-prompt consensus is the most
important operation within PVI.
\\
\textbf{Evidence for platform review.}
\sys returns a top-ranked carrier prompt together with its generated
images. A reviewer can therefore inspect both the lexical evidence and
the visual concept it elicits, even when the prompt does not reproduce
the original trigger. This output is designed to support moderation and
downstream investigation rather than make an enforcement decision by
itself. Failure to discover a carrier prompt is not evidence that an
adapter is benign or contains no undisclosed concept.

\section{Limitations}
\label{sec:limitations}

We evaluate separately loaded, U-Net--based LoRAs for SD1.5 and SDXL
under white-box access to the adapter and its compatible base model.
The current implementation searches short English-language, ASCII-only
Atomics at text-conditioned LoRA sites. The results therefore do not
establish performance for DiT-based generators, merged or quantized
adapters, black-box access, or multilingual and nonlexical carriers.

The formulation seeks one dominant hidden concept rather than an
exhaustive inventory. Adapters containing several comparably strong
concepts, or concepts deliberately distributed outside the inspected
subspaces, may require repeated or broader discovery. Evaluation uses
CLIP representations and one mean reference direction, verification
uses one controlled non-target per adapter, and the adaptive study
tests one fixed subspace-dispersion schedule. Broader safety concepts,
attacks, independent semantic encoders, human evaluation, and runtime
measurement remain important before deployment.

\section{Conclusion}
\label{sec:conclusion}

We formulate open-ended hidden-concept discovery for LoRA adapters and
develop \sys, which combines render-free parameter-guided search with
output-based visual identification. On the faithful SD1.5 cohort,
\sys reaches a mean target-concept discrepancy of 0.114, reducing
discrepancy by 70.2\% relative to the strongest baseline with the same
no-concept-input setting and by 37.7\% relative to reference-assisted
PEZ. Its discovered prompts further attain 0.980 precision, recall, and
F1 in controlled target-specificity evaluation. Under adaptive
subspace dispersion, conditional evasion remains 4.0\% among adapters
that preserve target-generation utility. More broadly, the evaluation
separates open-ended discovery from concept-specified verification.
\sys does not assume what a reviewer should test; it first extracts
parameter-supported candidates and then confirms them through generated
evidence. This matches the model-sharing setting, where the concealed
concept is unknown before inspection, and makes the returned
prompt--image pair directly useful for review rather than only reporting
a model-level score. These results demonstrate
that LoRA parameter structure can guide discovery of previously unknown
concepts and produce concrete prompts and images for subsequent human
review.

\par
\newpage
{\footnotesize \bibliographystyle{plain}
\bibliography{refs}}
\appendix
\raggedbottom

\section{Additional Details of SGES}
\label{app:sges-details}

\subsection{Text-Conditioning Site Selection}
\label{app:sges-sites}

SGES retains only LoRA sites whose inputs can be determined from text
conditioning without executing the diffusion denoising trajectory.
This allows LoRA-subspace alignment to be evaluated without image
generation.
\textbf{Retained sites and inputs.}
We retain LoRA sites in the text encoder and the key/value projections
of text cross-attention in the denoising backbone~\cite{rombach2022high,podell2024sdxl}.
For a text-encoder
site, $\mathbf h_{s,u}(x)$ is the contextual representation entering
that site when processing prompt $x$. For a retained cross-attention
site, $\mathbf h_{s,u}(x)$ is the final text representation supplied
by the text encoder. In both cases, the required inputs are available
from text conditioning alone. Query projections and other denoiser
LoRA sites are excluded because their inputs depend on the evolving
latent or subsequent denoising computation.
\textbf{Valid LoRA updates.}
For each selected site $s$, SGES requires both LoRA factors
$\mathbf A_s$ and $\mathbf B_s$ and forms the raw update
$\Delta\mathbf W_s=\mathbf B_s\mathbf A_s$. Sites with an incomplete
factor pair or an exactly zero update are discarded, while non-finite
updates are treated as invalid. For each retained site, the dominant
input directions are obtained from the SVD of the unscaled update
$\Delta\mathbf W_s$, retaining the top-$k$ right singular directions.
\textbf{Site grouping.}
Retained sites are grouped by their location in the model.
Text-encoder sites form the $\mathrm{te}$ group, while retained
cross-attention sites in the down and up paths of the denoising
backbone form the $\mathrm{down}$ and $\mathrm{up}$ groups,
respectively. Mid-block sites are excluded in our implementation.
Since an adapter may not contain retained sites in every group, we
define
\begin{equation}
\mathcal G_{\mathcal A}
=
\left\{
b\in\{\mathrm{te},\mathrm{down},\mathrm{up}\}
\;\middle|\;
|\mathcal J_b|>0
\right\}.
\label{eq:present-site-groups}
\end{equation}
Only groups in $\mathcal G_{\mathcal A}$ participate in subsequent
SGES computation.

\subsection{Composed-Word Discovery}
\label{app:bpe-beam}

A meaningful word may be split into multiple tokenizer units~\cite{sennrich2016neural,radford2021learning}, so
searching only individual vocabulary entries can miss strongly aligned
words. SGES therefore composes multiple lexical units into candidate
words before constructing the Atomic pools.
\textbf{Lexical units.}
We divide alphabetic tokenizer entries into prefix units and final
units according to the tokenizer's word-ending marker. Prefix units may
appear at the beginning or middle of a composed word, whereas final
units terminate the word. All units are scored separately for each site
group. Single-unit words are handled directly by the original
vocabulary and are therefore not produced by the composer.
\textbf{Progressive composition.}
SGES constructs words of two and three lexical units. For each group,
we first retain the top 30 prefix units and take their union across
groups. Each retained prefix is paired with every final unit to form
two-unit candidates. Final units are not pre-filtered by their
individual scores, allowing combinations whose alignment emerges only
after composition. All candidates are evaluated as complete text inputs
using Equations~\eqref{eq:shared-position}
and~\eqref{eq:composition-score}, and each group retains its top 120
two-unit candidates.

Three-unit candidates are formed by inserting a middle prefix into
promising two-unit compositions. The outer units are drawn from the
union of the top 40 two-unit candidates from each group, while the
middle-unit set contains the top 60 prefixes from each group together
with all prefixes of at most two characters. Each group retains its
top 80 three-unit candidates. Composition stops after three units.
\textbf{Validation and filtering.}
Retained compositions are decoded and accepted only if they remain unchanged after
tokenizer re-encoding and decoding. Candidates are deduplicated by
decoded text, with the shorter composition retained when multiple
chains produce the same word.

The valid words are then retokenized and rescored as complete
candidates. In each group, a composed word must score at least as high
as the strongest individual lexical unit. To suppress near-duplicate
compositions, let $P(w)$ denote the set of lexical units used to form
word $w$, and define
\begin{equation}
d_{\mathrm{piece}}(w,w')
=
\max\{|P(w)|,|P(w')|\}
-
|P(w)\cap P(w')|.
\label{eq:piece-family-distance}
\end{equation}
Candidates are processed in descending score and retained only if
$d_{\mathrm{piece}}(w,w')>1$ for every previously retained candidate.
We additionally allow each prefix or final unit to appear in at most
two retained compositions within a group.

After filtering, each group contributes at most 75 composed-word
proposals. Duplicate texts across groups are merged, and composed words
already represented by native whole-word vocabulary entries are
removed. The remaining proposals compete directly with native words
during Atomic pool construction in
Appendix~\ref{app:atomic-pool-details}.

\subsection{Atomic Candidate Pool Construction}
\label{app:atomic-pool-details}

SGES constructs a compact Atomic pool for each present site group from
two complementary candidate sources: lexical tokens and common
phrases.
\textbf{Candidate sources.}
We consider two types of Atomics. \textsc{Token} candidates consist of
native whole-word vocabulary entries and the composed words discovered
in Appendix~\ref{app:bpe-beam}. \textsc{Phrase} candidates are
drawn from the fixed public prompt corpus
Gustavosta/Stable-Diffusion-Prompts~\cite{gustavosta2026sdprompts}, shared across all
adapters and independent of target-specific information.

To construct the phrase source, corpus prompts are lowercased and
reduced to alphabetic words of at least two characters. We extract
adjacent word pairs occurring at least five times and retain those
tokenized into two or three lexical units. After removing duplicate
token sequences, the 3,000 most frequent phrases form the shared
candidate set. Corpus frequency is used only for source construction;
subsequent ranking depends solely on the LoRA-specific alignment score.
\textbf{Candidate selection.}
Each candidate is independently scored in every present group using
$R_b(a\mid a)$ from Equation~\eqref{eq:atomic-response}.
\textsc{Token} and \textsc{Phrase} candidates are ranked and filtered
separately. We remove non-ASCII candidates and retain only the
highest-scoring candidate within each lemma-equivalent family. We also
limit repeated lexical-root families to prevent closely related forms
from dominating a group pool.

Because the same Atomic may rank highly in multiple groups, each
candidate is assigned to the group in which it achieves the best
within-type rank. Let $\operatorname{rank}_b(a)$ denote this rank. We
define
\begin{equation}
\operatorname{owner}(a)
=
\operatorname*{arg\,min}_{b\in\mathcal G_{\mathcal A}}
\operatorname{rank}_b(a),
\label{eq:atomic-owner}
\end{equation}
with ties resolved by the fixed priority
$\mathrm{te}\succ\mathrm{up}\succ\mathrm{down}$.
\textbf{Final Atomic pools.}
Each present group retains the 30 highest-ranked \textsc{Token}
Atomics and the 20 highest-ranked \textsc{Phrase} Atomics assigned to that group:
\begin{equation}
\mathcal C_b
=
\mathcal C_b^{\mathrm{token}}
\cup
\mathcal C_b^{\mathrm{phrase}},
\quad
|\mathcal C_b^{\mathrm{token}}|=30,
\quad
|\mathcal C_b^{\mathrm{phrase}}|=20.
\label{eq:atomic-pool-composition}
\end{equation}
Native and composed words compete for the same \textsc{Token} quota,
so composed words receive no dedicated allocation. The resulting
group-specific pools provide the search space for evolutionary prompt
search.

\subsection{Evolutionary Search and Shortlist Construction}
\label{app:ea-shortlist}

Although the Atomic pools substantially reduce the lexical search
space, the number of prompt combinations remains large. Moreover,
$\operatorname{SAS}(p)$ depends on the jointly encoded prompt, so
Atomics cannot be optimized independently. SGES therefore performs
evolutionary search over complete prompts.
\textbf{Evolutionary search.}
Each present site group contributes three Atomic slots. Initial prompts
are sampled from the corresponding group-specific pools with a mixture
of rank-biased and uniform probabilities:
\begin{equation}
P(i)
=
0.3\frac{1}{n}
+
0.7
\frac{\exp(-r_i/\tau_{\mathrm{EA}})}
{\sum_j\exp(-r_j/\tau_{\mathrm{EA}})},
\qquad
\tau_{\mathrm{EA}}=100,
\label{eq:ea-rank-sampling}
\end{equation}
where $r_i$ is the zero-based rank of Atomic $i$ in a pool of size
$n$. Sampling excludes Atomic texts already used elsewhere in the same
prompt. Each run maintains a population of 48 prompts for 100
generations, and we perform three independent runs.

Prompts are ranked by contextual SAS\@. Parent selection uses three-way
tournaments~\cite{miller1995genetic}, followed by slot-wise crossover with probability $0.7$
and per-slot mutation. Both operations preserve the site-group
assignment of each slot and prevent duplicate Atomic texts. Fresh
random prompts are periodically injected to maintain exploration,
with a larger restart triggered when the best prompt remains unchanged
for three consecutive generations.
\textbf{Diversity preservation.}
Maximizing SAS alone can produce many minor variations of the same
high-scoring prompt. SGES therefore groups prompts according to their
Atomic-set overlap. Let $S(p)$ denote the set of Atomic texts in
prompt $p$. We define
\begin{equation}
d_{\mathrm{set}}(p,p')
=
\max\{|S(p)|,|S(p')|\}
-
|S(p)\cap S(p')|.
\label{eq:ea-family-distance}
\end{equation}
Prompts with $d_{\mathrm{set}}(p,p')\leq3$ are treated as belonging to
the same family, and only the highest-scoring member of each family
retains selection fitness~\cite{petrowski1996clearing}. The four strongest family representatives
are preserved as elites.
\textbf{Coordinate refinement.}
Every three generations, SGES further refines up to four of the highest-scoring family
representatives through coordinate-wise search. For each Atomic slot,
all valid replacements from its group-specific pool are evaluated while
the remaining slots are fixed. The replacement yielding the highest
contextual SAS is retained if it improves the current prompt. Up to two
passes are performed, with early termination when a complete pass
produces no improvement.
\textbf{Shortlist construction.}
After evolution, each run retains the five highest-scoring prompts as
\emph{confirm} candidates. To increase family diversity, SGES further
selects up to five \emph{reservoir} candidates from approximately the
top 5\% of all evaluated prompts, requiring a distance $d_{\mathrm{set}}>3$ from every previously selected candidate.

SGES additionally constructs shorter candidates from at most two
family-distinct high-scoring lineages. Starting from a full prompt,
Atomics are greedily removed according to contextual SAS while
preserving at least one Atomic from every originally present site
group. We retain candidates with lengths of 7, 5, and 3 whenever feasible.
Finally, candidates from all three runs are merged and deduplicated to
form
\begin{equation}
\mathcal P_{\mathrm{short}}
=
\{p_1,\ldots,p_N\},
\label{eq:sges-shortlist}
\end{equation}
which is passed to PVI for output-side concept identification.

\section{Evaluation Details}
\label{app:evaluation-details}

\subsection{Benchmark and Models}
\label{app:benchmark-details}

We evaluate our method on 164 LoRA adapters across the SD1.5 and SDXL
model families. The benchmark covers real identities, synthetic
identities, anime characters, and artistic styles. It combines 48
adapters collected from public model-sharing platforms~\cite{civitai2026models,huggingface2026modelhub} with 116
adapters trained locally under controlled settings.

The locally trained adapters focus on identity concepts, comprising
19 real identities and 10 synthetic identities. For each identity,
we train adapters under both SD1.5 and SDXL\@. Within each model family, we train
two matched variants that differ in how faithfully the target concept
is disclosed by the trigger. The \emph{faithful} variant uses a
human-readable trigger that reveals the target identity, whereas the
\emph{unfaithful} variant uses an opaque alphanumeric trigger that
does not reveal the target semantics. The matched variants use
identical training images and optimization settings and differ only in
the trigger used in their training captions. This construction yields 116 locally trained adapters.

Among the 164 adapters, 106 use faithful triggers and 58 use
unfaithful triggers. The latter comprise 29 opaque-trigger variants in
each model family. The baseline comparison, verification experiment,
and main ablations use one fixed faithfully disclosed SD1.5 cohort for
which every compared method is applicable. Cross-family and
trigger-disclosure results use their respective benchmark subsets, as
stated with each result.

Each adapter is evaluated with a compatible base model~\cite{rombach2022high,podell2024sdxl,stablediffusion2026v15modelcard,stabilityai2026sdxlmodelcard}. The benchmark
includes multiple compatible checkpoints from both the SD1.5 and SDXL
families.

\subsection{Local LoRA Training}
\label{app:controlled-training}

All locally trained identity adapters use LoRA rank $r=128$.
LoRA is applied to both the denoising network and the
text encoder. For SDXL, both the CLIP-L and OpenCLIP-G text encoders
are adapted~\cite{radford2021learning,cherti2023reproducible,podell2024sdxl,stabilityai2026sdxlmodelcard}. The main training configuration is summarized in
Table~\ref{tab:controlled-training}.

\begin{table}[H]
\centering
\caption{Training configuration for locally trained identity LoRAs.}
\label{tab:controlled-training}
\small
\begin{tabular}{lll}
\toprule
Setting & SD1.5 & SDXL \\
\midrule
Base model & Stable Diffusion 1.5 & SDXL 1.0 \\
Training resolution & 512 px & 1024 px \\
LoRA rank & 128 & 128 \\
U-Net learning rate & $1\times10^{-4}$ & $1\times10^{-4}$ \\
Text-encoder learning rate & $5\times10^{-5}$ & $5\times10^{-5}$ \\
Optimizer & AdamW & AdamW \\
LR schedule & Constant & Constant \\
Training steps & 3,000 & 1,000 \\
\bottomrule
\end{tabular}
\end{table}

We use AdamW~\cite{loshchilov2017decoupled} with $\beta=(0.9,0.999)$,
$\epsilon_{\mathrm{Adam}}=10^{-6}$, weight
decay $0.01$, and maximum gradient norm $1.0$. Training follows the
standard diffusion noise-prediction objective~\cite{ho2020denoising,rombach2022high}.
Given an image $x$, a training caption $p_{\mathrm{cap}}$,
a sampled timestep $t$, and Gaussian noise $\epsilon$, the model minimizes
\begin{equation}
\mathcal L_{\mathrm{train}}
=
\mathbb E_{x,p_{\mathrm{cap}},t,\epsilon}
\left[
\left\|
\epsilon-
\epsilon_{\theta}(z_t,t,p_{\mathrm{cap}})
\right\|_2^2
\right].
\label{eq:lora-training-loss}
\end{equation}

All SD1.5 adapters are trained for 3,000 optimization steps, whereas
all SDXL adapters are trained for 1,000 steps.

\subsection{Training Data for Locally Trained LoRAs}
\label{app:identity-data}

The locally trained LoRAs use two sources of identity images. Real
identities are represented by web-collected photographs of public
figures, whereas synthetic identities are represented by images
generated from fixed synthetic identity specifications. Both sources
follow the same captioning protocol for LoRA training.
\textbf{Real-identity images.}
For each of the 19 real identities, we collect publicly available
images of the corresponding public figure through web image search.
The resulting training sets contain 19--46 images per identity and
cover varied viewpoints, poses, clothing, scenes, and photographic
conditions.
\textbf{Synthetic-identity images.}
For each of the 10 synthetic identities, we first construct a synthetic
identity specification describing a fixed set of identity-defining
visual attributes, including face shape, eyes, hair, eyebrows, nose,
jaw structure, and a distinctive feature. GPT-5.5 via Codex~\cite{openai2026gpt55} is used to generate specifications.

Given an identity specification, we then generate a batch of images
while keeping its identity-defining attributes fixed and varying only
image-specific factors such as viewpoint, pose, expression, clothing,
scene, lighting, framing, and shot scale. The same specification is
used throughout the batch to maintain identity consistency across
images. Synthetic training sets contain 25--50 images per identity.
\textbf{Training captions.}
Real and synthetic images follow the same captioning protocol. For each
training image, GPT-5.5 via Codex~\cite{openai2026gpt55} generates an image-specific caption
describing visible factors that vary across examples, such as framing,
viewpoint, pose, expression, clothing, scene, lighting, and occlusion.

Each caption begins with the trigger and a class word, uses the trigger
exactly once, and contains 20--45 English words. For each matched
faithful and unfaithful LoRA pair, the images and caption content are
held fixed; only the identity trigger is replaced. The faithful variant
uses the human-readable identity name, whereas the unfaithful variant
uses the corresponding opaque alphanumeric trigger.
\textbf{Generation and captioning templates.}
The principal templates used for synthetic-image generation and
training-caption construction are provided below. Placeholders in
braces are instantiated from the fixed identity specification or
trusted training metadata.

\begin{promptbox}{Synthetic identity specification}
Design {N} entirely synthetic human identities for a controlled
diffusion-model benchmark. None should resemble or be named after a
real public figure.

Make the identities visually distinguishable using independent
combinations of age band, face shape, skin tone, eye shape and color,
hair texture, color, and cut, eyebrow geometry, nose geometry, jaw or
chin geometry, and one subtle distinctive feature.

Keep the attributes of each identity internally consistent. Avoid
occupations, personality descriptions, stereotypes, costumes, fantasy
features, or exaggerated deformities.

Return JSON containing:
id, synthetic_name, class_word, age_band, face_shape, skin_tone, eyes,
hair, eyebrows, nose, jaw_or_chin, distinctive_feature, and
identity_signature.
\end{promptbox}

\begin{promptbox}{Synthetic identity generation}
Generate a diverse batch of photorealistic images of the same
synthetic identity defined below.

IDENTITY:
{IDENTITY_SPECIFICATION}

The identity-defining attributes in the specification must remain
consistent across all generated images. Every image must depict the
same person.

Across the batch, vary only image-specific factors, including:
shot scale, viewpoint, pose, gaze, expression, clothing, scene,
lighting, framing, and aspect ratio.

Include diverse close-up, head-and-shoulders, upper-body,
three-quarter, and full-body views; front, three-quarter, and profile
viewpoints; indoor and outdoor scenes; and varied natural photographic
conditions.

Maintain realistic human anatomy, natural skin texture, and consistent
facial identity across the entire batch. Do not introduce additional
people, text, watermarks, or logos, and do not make any image resemble a real
public figure.@*)
\end{promptbox}
\begin{promptbox}{Training caption generation}
You are a conservative visual annotator creating captions for an
identity-conditioned LoRA training dataset. Describe only directly
visible, image-specific factors that vary across images. Do not infer
the person's identity from the image and do not repeatedly describe
stable identity-defining facial attributes.

TRIGGER: "{TRIGGER}"
CLASS_WORD: "{CLASS_WORD}"

Generate a (*@20--45-word@*) English caption that:
1. begins with the exact TRIGGER followed by CLASS_WORD;
2. uses TRIGGER exactly once;
3. describes only visible factors such as shot scale, viewpoint, pose,
   gaze, expression, clothing, scene, lighting, occlusion, and
   rendering;
4. contains no additional proper name or inferred identity.
\end{promptbox}

\subsection{Baseline Configurations}
\label{app:baseline-details}

We compare with six prior methods from two related problem settings:
concept-agnostic difference discovery and concept-specified
verification. Table~\ref{tab:baseline-overview} summarizes how each
method is used in our LoRA setting and the concept information it
requires. We use official implementations when available~\cite{driftscope2026code,compcon2026code,visdiff2026code,dpodiff2026code,pez2026code}.Since PAIA
does not provide public code, we independently reproduce it from the
published method.

\begin{table}[H]
\centering
\caption{Prior methods, their use in our LoRA evaluation, and the
concept information they require.}
\label{tab:baseline-overview}
\small
\begin{tabular}{@{}>{\raggedright\arraybackslash}p{0.21\columnwidth}
                    >{\raggedright\arraybackslash}p{0.42\columnwidth}
                    >{\raggedright\arraybackslash}p{0.24\columnwidth}@{}}
\toprule
Method & Use in LoRA evaluation & Required input \\
\midrule
DriftScope
& Rank tokens by their LoRA--base drift
& None \\

CompCon
& Search for prompts exposing LoRA--base differences
& None \\

VisDiff
& Describe differences between paired LoRA and base image sets
& None \\

DPO-Diff-EPO
& Optimize prompts for paired-model divergence
& None \\
\midrule

PEZ
& Recover a hard prompt from concept images
& Concept images \\

% UnlearnDiffAtk
% & Known-concept prompt recovery
% & \makecell[l]{Recover prompts through concept-image\\reconstruction on the LoRA model}
% & Concept images \\
% \midrule

PAIA
& Verify supplied concepts for each LoRA
& Candidate images \\
\bottomrule
\end{tabular}
\end{table}
For every prompt-returning baseline, the reported carrier is selected
using that method's own score or ranking; target-concept references are
used only for final evaluation and never for carrier selection.
We
retain each method's standard search configuration and use the same
generation and final-evaluation protocol wherever the method's task
permits. PEZ receives a concept-image set that is disjoint from the
held-out references used to compute target-concept discrepancy.
\textbf{Concept-agnostic difference discovery.}
These methods do not assume a concept to be examined in advance.
Instead, we adapt them to expose changes introduced by the LoRA through
model comparison, output comparison, or prompt optimization.

\emph{DriftScope}~\cite{laria2026driftscope} studies semantic drift
introduced by diffusion-model adaptation. Given an original and an
adapted model, it identifies lexical tokens whose associated visual
concepts change most after adaptation. We instantiate the model pair
with the compatible base and LoRA-enabled models and insert the
resulting high-drift tokens into DriftScope's prompt templates to form
candidate prompts.

\emph{CompCon}~\cite{dunlap2025discovering} compares two T2I models and
discovers visual attributes on which their generations diverge,
together with prompts that expose these differences. We use the
LoRA-enabled and compatible base models as the model pair. Starting
from generic prompts, CompCon identifies LoRA--base visual differences
and searches for prompts that reliably expose them. The resulting
verified prompts are used as candidate prompts.

\emph{VisDiff}~\cite{dunlap2024describing} describes visual differences
between two fixed image sets. We construct one set from LoRA-enabled
generations and the other from compatible base-model generations using the same
generic prompts and random seeds. VisDiff produces natural-language
descriptions that best distinguish the two sets, which we directly use
as candidate prompts. Unlike CompCon, VisDiff describes differences in
fixed image sets rather than searching for new prompts that expose
them.

\emph{DPO-Diff-EPO}~\cite{wang2024discrete} performs discrete
prompt optimization for diffusion models. We retain its EPO optimizer
but replace the original task objective with one that favors visual
divergence between paired LoRA-enabled and compatible base-model generations.
The resulting prompts therefore expose generation behavior changed by
the LoRA.

\textbf{Concept-specified verification.}
These methods require the concept to be specified in advance. PEZ
recovers prompts from concept examples and provides
indirect evidence of whether the concept can be elicited, whereas PAIA
directly verifies whether the specified concept is encoded.

\emph{PEZ}~\cite{wen2023hard} performs image-guided hard-prompt
optimization. Given concept images, it optimizes a prompt in continuous
embedding space and projects it onto discrete vocabulary tokens. In our evaluation, PEZ accesses neither the
LoRA adapter under evaluation, its compatible base model, nor its
training trigger; it receives only the supplied concept images. The
recovered prompt is subsequently evaluated on the LoRA-enabled model to
determine whether it elicits the specified concept.

% \emph{UnlearnDiffAtk}~\cite{zhang2024unlearndiffatk} originally
% generates adversarial prompts that recover a known concept from a
% concept-unlearned diffusion model. In our evaluation, we replace the
% unlearned model with the LoRA-enabled model and apply its prompt
% optimization to each specified concept. Each concept image is paired
% with an automatically generated caption, from which additional prompt
% tokens are optimized. The resulting prompts are ranked by their
% agreement with the concept-image set.

\emph{PAIA}~\cite{yuan2025lurks} is the closest prior work in terms of problem
setting. It verifies whether a LoRA-adapted diffusion model has learned
a prespecified concept by comparing its denoising behavior with that of
the compatible base model. Given images of the candidate and irrelevant concepts,
it returns a binary learned/not-learned
decision. We query each candidate concept independently using its
concept images. PAIA therefore verifies supplied concepts rather than
discovering an unknown concept or producing a carrier prompt. Since no
official implementation is publicly available, we independently
reproduce the published method.

\section{Additional Sensitivity Analysis}
\label{app:additional-ablations}

\subsection{Retained Subspace Dimension}
\label{app:ablation-k}

Every value of $k$ reconstructs the Atomic pools and reruns the complete
discovery pipeline rather than reusing candidates from the default
configuration.

\begin{table}[H]
\centering
\caption{Effect of the maximum number of leading LoRA input directions retained per site.
Lower target-concept discrepancy is better.}
\label{tab:ablation-k}
\small
\setlength{\tabcolsep}{4pt}
\begin{tabular}{@{}ccccc@{}}
\toprule
$k$ &
Real ID $\downarrow$ &
Synthetic ID $\downarrow$ &
Anime $\downarrow$ &
Mean $\downarrow$ \\
\midrule
1
& 0.080 & 0.202 & 0.144 & 0.128 \\
\textbf{2}
& \textbf{0.078}
& \textbf{0.178}
& \textbf{0.123}
& \textbf{0.114} \\
4
& 0.088 & 0.181 & 0.144 & 0.127 \\
\bottomrule
\end{tabular}
\end{table}

For site $s$, the number of retained directions is
$k_s=\min(k,\rho_s)$, where
$\rho_s=\operatorname{rank}(\Delta\mathbf W_s)$. Table~\ref{tab:ablation-k}
varies $k$ while selecting the
leading directions. Retaining two directions performs best, and the
mean varies by at most 0.014 over the tested values. With
$k=1$, real-identity performance is nearly unchanged (0.080 versus
0.078), whereas synthetic identities and anime characters degrade more
noticeably. Increasing to $k=4$ primarily hurts real identities and
anime. This pattern is consistent with one direction sometimes omitting
complementary responses and four directions admitting weaker ones. The
experiment therefore shows limited sensitivity within
$k\in\{1,2,4\}$, but does not establish a universal optimum outside
that range.

\section{Adaptive Attack Details}
\label{app:adaptive-attacks}

\subsection{Threat Model and Controls}
\label{app:adaptive-threat}

The adaptive creator knows that \sys uses leading LoRA input
directions, but not the number of retained directions, the selected
site groups, or the remaining hyperparameters used by \sys. The creator can modify LoRA training but must preserve the
LoRA-enabled model's ability to generate the adapter's dominant hidden concept. In this controlled
evaluation, utility is measured against the target concept $c$.

We apply the attack to the controlled faithful SD1.5 identity
adapters. Each attacked adapter uses exactly the same training images,
captions, human-readable trigger, compatible base model, LoRA target
modules, rank and scaling factor, optimization steps, optimizer, learning rates,
scheduler, batch configuration, resolution, and seed policy as its
matched clean adapter. No LoRA module is added, removed, or moved.
The sole change is the regularizer described below, applied to every
LoRA module already trained by the original recipe.

\subsection{Subspace-Dispersion Objective}
\label{app:adaptive-objective}

For a nonzero trained LoRA module $s$, let
\begin{equation}
\Delta\mathbf W_s=\mathbf B_s\mathbf A_s
\end{equation}
have singular values $\{\sigma_{s,j}\}$. We define its scale-invariant
spectral concentration as
\begin{equation}
\operatorname{SC}_s
=
\frac{\sum_j\sigma_{s,j}^4}
{\left(\sum_j\sigma_{s,j}^2\right)^2}.
\label{eq:module-spectral-concentration}
\end{equation}
A spectrum concentrated in a few directions has a larger
$\operatorname{SC}_s$, whereas a flatter spectrum has a smaller value.
Let $\mathcal M_{\mathrm{LoRA}}$ contain every LoRA module in the clean
training recipe. The attack minimizes
\begin{equation}
\mathcal R_{\mathrm{disp}}
=
\frac{1}{|\mathcal M_{\mathrm{LoRA}}|}
\sum_{s\in\mathcal M_{\mathrm{LoRA}}}\operatorname{SC}_s.
\label{eq:dispersion-regularizer}
\end{equation}
The spectral concentration reported in
Section~\ref{sec:adaptive-attacks} is this module-wise mean.

The regularizer is computed without forming the full update matrix.
Let
\begin{equation}
\mathbf G_{A,s}=\mathbf A_s\mathbf A_s^\top,
\qquad
\mathbf G_{B,s}=\mathbf B_s^\top\mathbf B_s.
\end{equation}
Then the same concentration can be computed exactly as
\begin{equation}
\operatorname{SC}_s
=
\frac{
\operatorname{tr}
(\mathbf G_{A,s}\mathbf G_{B,s}
 \mathbf G_{A,s}\mathbf G_{B,s})
}{
\left[\operatorname{tr}
(\mathbf G_{A,s}\mathbf G_{B,s})\right]^2
}.
\label{eq:efficient-spectral-concentration}
\end{equation}
This computation uses only rank-dimensional matrices and preserves
gradients. The Gram matrices, concentration, and dispersion loss are
computed in FP32; the original mixed-precision diffusion training is
otherwise unchanged.

\subsection{Fixed Training Schedule}
\label{app:adaptive-schedule}

Let $\nu$ denote the optimization step. The attacked training
objective is
\begin{equation}
\mathcal L_{\mathrm{atk}}(\nu)
=
\mathcal L_{\mathrm{train}}
+
\lambda_{\mathrm{disp}}(\nu)\mathcal R_{\mathrm{disp}},
\label{eq:dispersion-training-objective}
\end{equation}
where the attack strength is fixed rather than selected using \sys's
discovery results:
\begin{equation}
\lambda_{\mathrm{disp}}(\nu)
=
\begin{cases}
0, & \nu<300,\\
0.1(\nu-300)/300, & 300\leq \nu<600,\\
0.1, & \nu\geq600.
\end{cases}
\label{eq:dispersion-schedule}
\end{equation}
The initial warm-up avoids evaluating the concentration ratio when the
LoRA update is near zero. The coefficient is not searched or adjusted
after training.

\subsection{Evaluation Protocol}
\label{app:adaptive-protocol}

Section~\ref{sec:adaptive-attacks} defines utility preservation and
conditional evasion. Each matched clean--attacked pair uses the same
fixed $p_{\mathrm{trig}}$ containing the human-readable training
trigger; it is set before attack training and unavailable to the
moderator. We render $p_{\mathrm{trig}}$ and the carrier prompt $p^\star$
with the attacked LoRA-enabled model $\mathcal B_{\mathcal A}$ and
evaluate both against the same attack-independent held-out references
for the target concept $c$. Training uses neither the references nor
discovery results.

\end{document}